\documentclass[12pt]{article}
\usepackage[utf8]{inputenc}

\usepackage{amssymb}
\usepackage{booktabs}
\usepackage{graphicx}
\usepackage{subfig}

\makeatletter
\newif\if@blind 
\@blindfalse 

\usepackage[preprint]{style}
\usepackage{fancyhdr}

\begin{document}


\title{Two Sparsities Are Better Than One: Unlocking the Performance Benefits of Sparse-Sparse Networks}

    \if@blind
\author{Authorship removed for blind review}
\else
\author{Kevin Hunter, Lawrence Spracklen and Subutai Ahmad \\
Numenta, Redwood City, CA, USA \\
\texttt{\{khunter,\,lspracklen,\,sahmad\}@numenta.com}}
\fi

\pagestyle{fancy}
\lhead{Hunter \emph{et al.}}
\rhead{Two Sparsities Are Better Than One}

\maketitle

\begin{abstract}
In principle, sparse neural networks should be significantly more efficient than traditional dense networks. Neurons in the brain exhibit two types of sparsity; they are sparsely interconnected and sparsely active. These two types of sparsity, called weight sparsity and activation sparsity, when combined, offer the potential to reduce the computational cost of neural networks by two orders of magnitude. Despite this potential, today’s neural networks deliver only modest performance benefits using just weight sparsity, because traditional computing hardware cannot efficiently process sparse networks. In this article we introduce Complementary Sparsity, a novel technique that significantly improves the performance of dual sparse networks on existing hardware. We demonstrate that we can achieve high performance running weight-sparse networks, and we can multiply those speedups by incorporating activation sparsity. Using Complementary Sparsity, we show up to 100X improvement in throughput and energy efficiency performing inference on FPGAs. We analyze scalability and resource tradeoffs for a variety of kernels typical of commercial convolutional networks such as ResNet-50 and MobileNetV2. Our results with Complementary Sparsity suggest that weight plus activation sparsity can be a potent combination for efficiently scaling future AI models.

\end{abstract}


\section{Introduction}
In recent years, larger and more complex deep neural networks (DNNs) have led to significant advances in artificial intelligence (AI). However, the exponential growth of these models threatens forward progress. Training requires large numbers of GPUs or TPUs, and can take days or even weeks, resulting in large carbon footprints and spiraling cloud costs~\citep{carbon, Thomson2020}.

\begin{figure}
    \centering
    \includegraphics[width=0.8\textwidth]{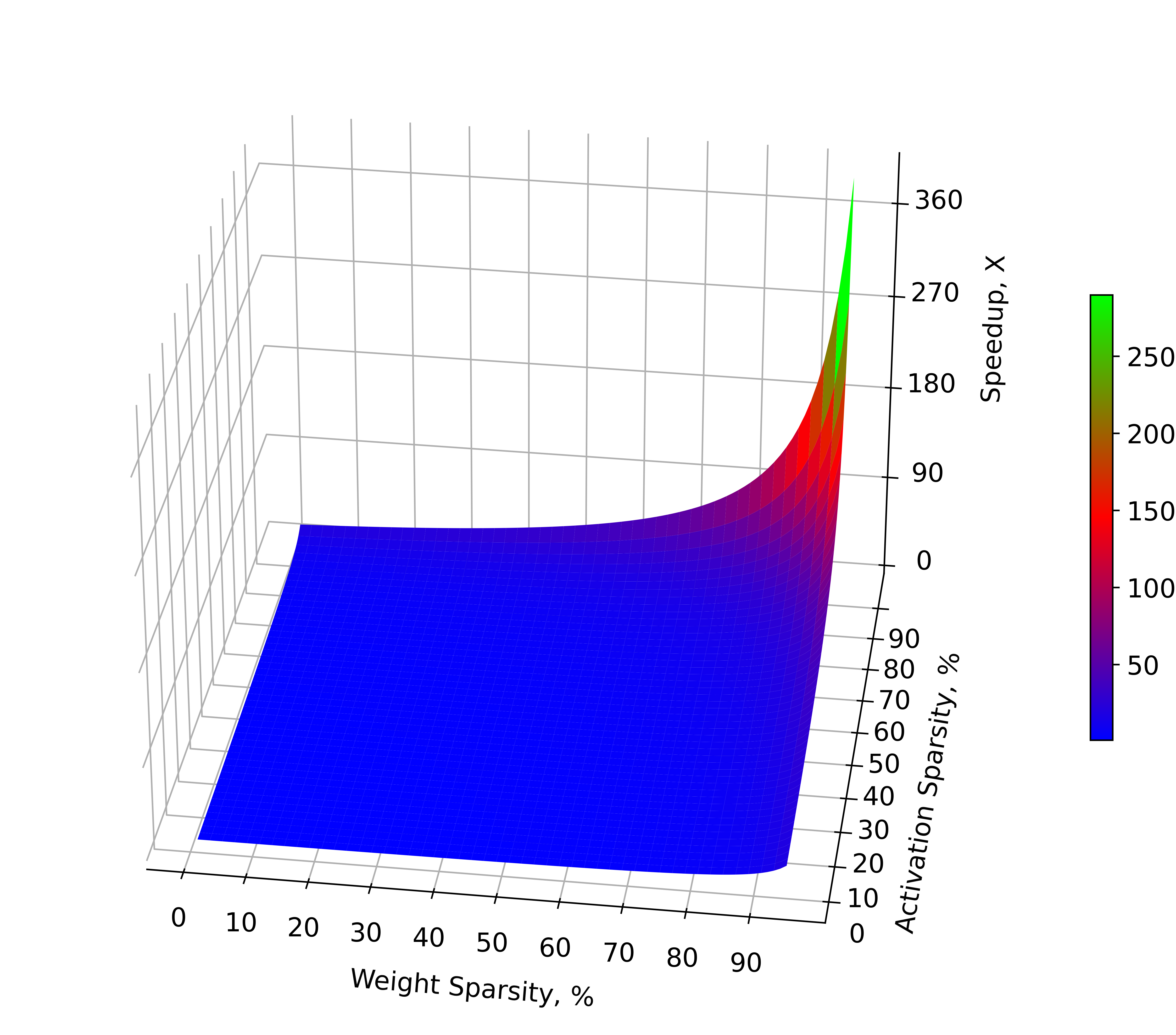}
    \caption{An illustration of the potential speedups that can be achieved with sparse networks (compared to their dense equivalents). When weight and activation sparsity are leveraged simultaneously (a sparse-sparse network), the benefits are multiplicative, enabling speedups that exceed two orders of magnitude.}
    \label{fig:multiplicative}
\end{figure}

Taking inspiration from neuroscience, sparsity has been proposed as a solution to this rapid growth in model size. Sparse networks either constrain the connectivity (weight sparsity) or activity (activation sparsity) of their neurons, significantly reducing both the size and computational complexity of the model. Typically, these techniques are applied in isolation to create \textit{sparse-dense} networks. However, weight and activation sparsity are synergistic, and when deployed in combination, the computational savings are multiplicative. Consequently, \textit{sparse-sparse} networks have the potential to reduce the computational complexity of the model by over two-orders of magnitude. For example, as illustrated in Figure \ref{fig:multiplicative}, when a network is 90\% weight sparse, only 1 out of every 10 weights is non-zero, facilitating a 10-fold reduction in compute. When a network is 90\% activation sparse, only 1 out of every 10 inputs is non-zero, similarly delivering a 10-fold reduction in compute. When applied in concert, the zero-values interplay, such that on average only 1 out of every 100 results will be non-zero, delivering a 100-fold savings, if techniques can be developed to efficiently skip fetching, multiplying and storing zero valued elements.

However, with current implementations, the resulting speedups only represent a small fraction of these theoretical computational savings~\citep{Gale2020}. The irregular patterns of neuron interconnections and activity introduced by sparsity have proved difficult to exploit on modern hardware, which is optimized for the execution of regular dense data structures. This ``impedance mismatch" significantly erodes the performance of sparse-dense networks and impedes the implementation of efficient sparse-sparse networks. Hardware platforms with dedicated logic for exploiting sparsity have begun to appear~\citep{ai_survey}, but the performance gains remain modest~\citep{ampere}.

In this article we discuss Complementary Sparsity, a novel solution that inverts the sparsity problem. Rather than creating hardware to support unstructured sparse networks, we illustrate how sparsity can be structured to match the requirements of the target hardware. We demonstrate that this solution both creates highly efficient weight-sparse networks, and establishes viable sparse-sparse networks, yielding large multiplicative benefits.

We investigate the potential of Complementary Sparsity and sparse-sparse networks on FPGAs, due to their flexible architecture. This flexibility provides an ideal laboratory for investigating the trade-offs associated with different implementation approaches, and enables us to refine our understanding of sparse-sparse resource requirements. The resulting implementations not only provide a path to highly efficient sparse-sparse network inference on FPGAs, but also provide insights that can be leveraged as IP blocks in other architectures or ASICs, or adapted to fit a wide range of other compute architectures.

In this paper, we make four main contributions:
\begin{enumerate}
    \item We introduce Complementary Sparsity, a novel form of structured sparsity.
    \item We establish how Complementary Sparsity can enable the construction of efficient sparse-sparse networks.
    \item We discuss our sparse-sparse network implementation on a FPGA, demonstrating a 110X speedup over an optimized dense implementation.
    \item We demonstrate that leveraging activation sparsity reduces the hardware resource utilization associated with the core components of convolutional networks.
\end{enumerate}

\section{Sparsity in the Brain, in Deep Learning, and in Hardware}
\label{sec:sparse}

\subsection{Sparsity in the Brain}
It is well known that neural activity in the brain, specifically the neocortex, is highly sparse. However, sparsity in the neocortex is instantiated in multiple different ways. Numerous studies show that only a small percentage of neurons become active in response to sensory stimuli \citep{Attwell2001, Weliky2003, Barth2012}. On average less than $2\%$ of neurons fire for any given input. This is true for all sensory modalities as well as areas that deal with language, abstract thought, planning, etc. Sparsity is also present in neural connectivity. Cortical pyramidal neurons show highly sparse connectivity to each other and receive relatively few excitatory inputs from most surrounding neurons \citep{Holmgren2003}. The density of local area connections is often less than $5\%$ \citep{Holmgren2003}. The brain is incredibly power efficient, a fact that has been directly linked to both activation sparsity \citep{Attwell2001, Lennie2003TheComputation} and connection sparsity \citep{Pulido2021SynapticTerminals}. Sparsity has also been linked to the brain's ability to form useful representations \citep{Olshausen1996, Olshausen2004}, make predictions \citep{Hawkins2016, Vinje2000, Miller2014}, as well as detect surprise and anomalies. It seems evident that sparsity is ubiquitous in the neocortex and fundamental to its efficiency and functionality.

\begin{figure}
    \centering
    \includegraphics[width=5.5in]{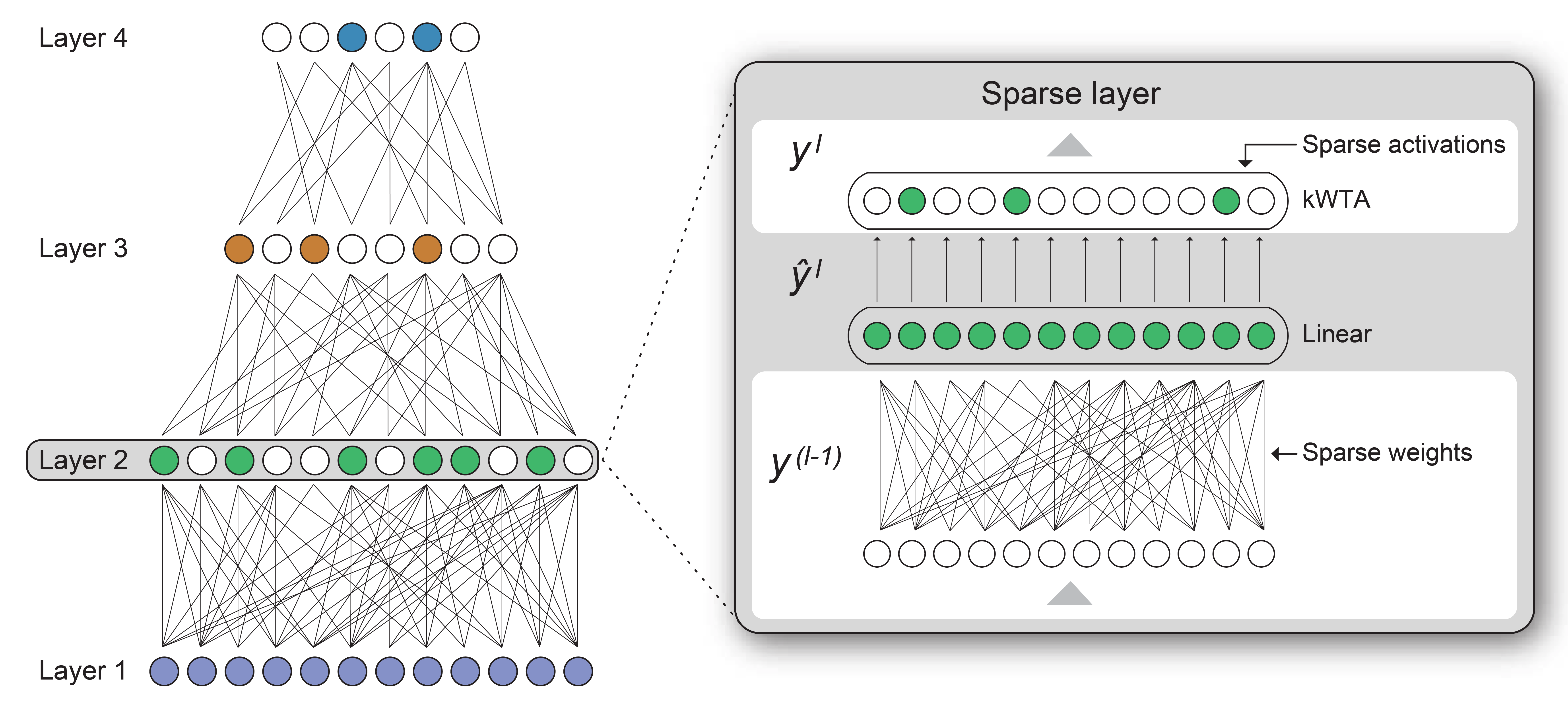}
    \caption{We focus on sparse networks with both sparse connectivity and sparse activations. Zeroes in the weight matrices enforce sparse connectivity. A $k$-WTA activation function ensures activations with fixed sparsity.}
    \label{fig:sparsenet}
\end{figure}

\subsection{Sparsity in Deep Learning}
\label{sec:dnn_sparsity}
The prevalence of sparsity in the brain stands in contrast to standard DNNs where sparsity is still a research area. DNNs are composed of stacked layers of linear neurons, with the output of layer \textit{N} forming the input to layer \textit{N+1}. Neurons compute a weighted sum of their inputs, with the weights for the neurons in each layer typically being expressed as a 2D weight matrix. The output of a layer can be computed as a simple matrix multiplication; the inputs to the layer form either an activation vector (in the case of a single input), or a matrix (when a batch of inputs are being processed). In standard DNNs both the weight matrices and the activation vectors are dense.

Analogous to the neurology, it is possible to create two forms of sparsity in DNNs: sparse connections and sparse activations (Figure~\ref{fig:sparsenet}). Absent connections are represented by 0's in the weight matrices, while inactive neurons are represented by 0's in the inputs to each layer. Recently, there has been an increase in research focused on creating networks that are both sparse and accurate. A variety of techniques have been proposed in the literature to achieve either form of sparsity, as summarized in the following sections. Note however that, unlike biology, it is rare for DNNs to combine both types of sparsity in the same network. 

\subsubsection{Weight Sparsity}
In contemporary dense networks, neurons are fully connected, with each neuron in a layer receiving input from each and every neuron in the previous layer. Research has shown that many DNNs are heavily overparameterized, and sparsity can be successfully applied to these networks \citep{Lee2008,Chen2018b}. This technique of limiting neuron interconnectivity is referred to as \emph{weight sparsity}.

In weight-sparse networks, each neuron is only connected to a subset of the neurons in the prior layer. As the sparsity of the weight matrices is increased the overall accuracy can drop. However, a variety of techniques have been developed to create networks that are both sparse and accurate \citep{sparsity-review}. Most research has focused on the creation of: a) sparse models by direct training; or, b) sparse models from existing dense networks by removing (or ‘pruning’) the least important weights \citep{sparsity-review}. Within these two broad approaches exist a variety of different techniques, with varying degrees of sophistication and dynamism. Most simplistic are single-shot pruning algorithms \citep{sparsity-review}, that remove all of the weights necessary to achieve the desired sparsity in one event. Iterative algorithms gradually increase the sparsity over the span of a number of steps until the desired sparsity is achieved \citep{sparsity-review}. In addition to undertaking iterative pruning, algorithms can iteratively grow connections, working to ensure that the optimal set of interconnections is retained \citep{pmlr-v119-evci20a}.

Pruning techniques primarily focused on reducing computational overheads are also in use \citep{Mocanu2018, pmlr-v119-evci20a}. Different levels of sparsity in each layer allows sparsity to be focused in components of the model to deliver the most significant speedups, while smaller layers that only contribute minimally to the overall computational costs and parameter counts are protected. Novel pruning techniques and sparsity patterns that are tailored to hardware requirements have also focused on convolutional layers, where the multiplicity of channels provides opportunity for a variety of hardware friendly sparsity patterns \citep{Changpinyo2017}.

\begin{figure}[b]
    \centering
    \includegraphics[width=13cm]{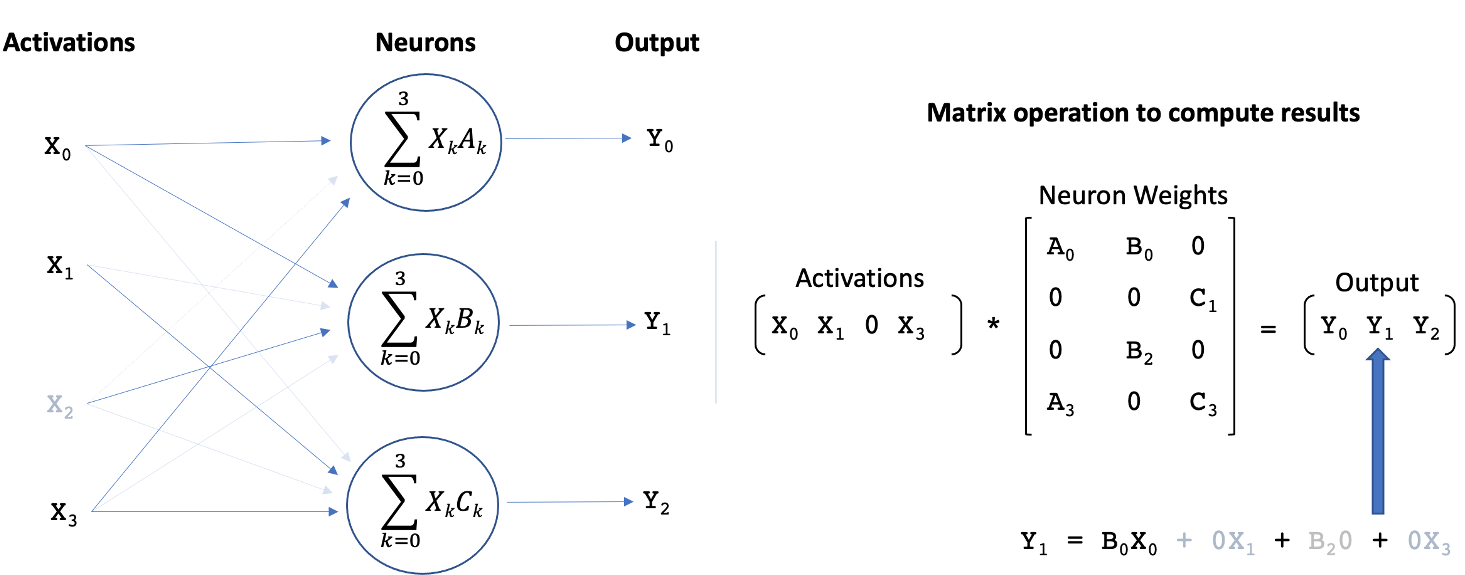}
    \caption{An illustration illustrating the simplification of matrix multiplication operations when both activations and weights are sparse. It is only necessary to compute a product if it contains a non-zero element in both the input activation \emph{and} the weight matrix.}
    \label{fig:multiply_zero}
\end{figure}

\subsubsection{Activation Sparsity}
\label{sec:activation_sparsity}
In DNNs the outputs (activations) of each layer are generally dense, with between $50\%$ to $100\%$ of the neurons having non-zero activations. While less commonly discussed, activation sparsity can also be applied to DNNs~\citep{Makhzani2015,Ahmad2019,thres-relu,10.1167/jov.20.12.10}. For activation sparsity, the determination of neurons to activate is typically performed by either explicitly selecting the top-k activations (frequently termed $k$-Winner-Take-All ($k$-WTA)) ~\citep{Makhzani2015,Ahmad2019} or by computing dataset specific activation thresholds for the neurons that, on average, reduce the number of activated neurons to the desired level~\citep{thres-relu}. It is also possible to reduce the overall magnitudes of activations by introducing appropriate regularizers that penalize large values \citep{Olshausen1996, Lee2008}.

In this article we focus on networks using $k$-WTA \citep{majani1989k, Cui2017} (Figure~\ref{fig:sparsenet}). In these networks the ReLU activation function is replaced by an activation function where the output of each layer is constrained such that only the $K$ most active neurons are allowed to be non-zero~\citep{Makhzani2015,Ahmad2019}. Whereas ReLU allows all activations above $0$ to propagate, $k$-WTA allows exactly the top $K$ activations to propagate. $k$-WTA thus provides the ability to precisely control the percentage of active neurons by setting $k$ appropriately. Typically $K$ is much smaller than the number of neurons in that layer. A detailed description of the activation function and results on several datasets can be found in~\citep{Ahmad2019}.

\subsection{Sparse DNNs in Hardware}
\label{sec:hw_bad}

\subsubsection{Theoretical Performance Benefits of Sparsity}
Sparse models provide a variety of potential benefits compared to traditional dense models, including model compression and computational efficiency. In weight-sparse networks, each absent connection is represented by a zero-valued element in the weight matrix, which in turn translates into a reduction in the number of multiply-accumulate (MAC) operations required for matrix multiplication. Accordingly, the theoretical computational savings associated with weight sparsity are directly proportional to the degree of weight sparsity. 

Activation sparsity can also deliver significant computational savings. When an incoming activation is zero valued, its contributions to all of the neurons' weighted sums will also be zero, allowing the computations of these sub-products to be skipped. Similar to the computational savings associated with weight sparsity, the computational savings from activation sparsity are directly proportional to the degree of activation sparsity.

Activation and weight sparsity can be applied to DNNs independently or in combination. With both forms of sparsity independently reducing the number of MAC operations required, exploiting both forms of sparsity has the potential to yield significant computational savings (Figure~\ref{fig:multiplicative}). MAC operations can be eliminated if either the corresponding input or the corresponding weight is zero (Figure~\ref{fig:multiply_zero}). These savings can be realized if the underlying hardware is capable of efficiently performing operations on these sparse matrices.

In practice it has proved extremely difficult to realize these performance benefits on current hardware architectures. Even for DNNs with high-degrees of weight sparsity, the performance gains observed are small. For example, on CPUs, even for weight sparse networks in which 95\% of the neuron weights have been eliminated, the performance improvements observed are typically less than 4X~\citep{nm_slow}. In addition, there are almost no techniques that simultaneously exploit both weight and activation sparsity. In part due to these difficulties, sparse networks have not been widely deployed in commercial settings. 

\begin{figure}[b]
    \centering
    \includegraphics[width=10cm]{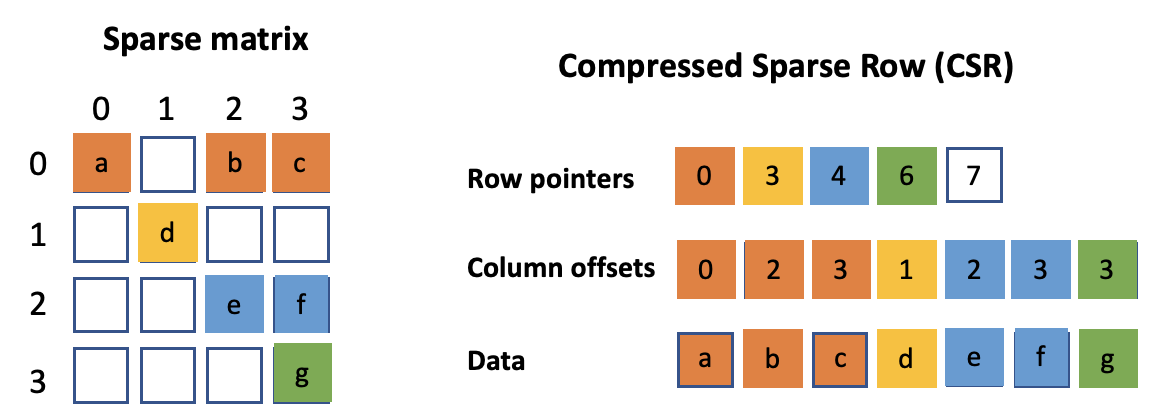}
    \caption{The Compressed Sparse Row (CSR) format for representing sparse matrices provides a compact representation, but requires index arrays to interpret the data.}
    \label{fig:csr}
\end{figure}

\subsubsection{Challenges of Accelerating Sparse Matrices}
Modern hardware architectures thrive on processing dense, regular data structures. Dense DNNs are perfectly suited to these architectures; their regular-dense matrix operations are painstakingly optimized for peak performance on every conceivable hardware platform.

For sparse models, performance gains are possible if the underlying hardware is capable of efficiently performing sparse matrix operations, skipping the unnecessary computations associated with the zero valued elements. The efficient rendezvous of non-zero weights with non-zero activations ensures the minimum number of operations, enabling sparse-sparse networks to reduce computations and the attendant power consumption.

Connections between neurons in sparse networks are ideally determined to maximize the accuracy of the network. As such, there is typically no regular pattern to these connections. Similarly, the locations of the non-zero activations will likely display no regular pattern, and will change constantly. The resulting irregular patterns of non-zero elements in the weight and activation matrices present challenges to efficiently exploiting sparsity on today's hardware architectures.

Sparse matrices are often represented in a compressed form, where only the non-zero elements are retained, along with sufficient indexing information to locate the elements within the matrix, as illustrated in Figure~\ref{fig:csr}, or are highly structured (such as the Band matrices and Block Diagonal Matrices~\citep{hpc}. These are often found in the context of High Performance Computing (HPC) software packages~\citep{bank1992sparse}. However, given the overheads associated with explicitly storing and retrieving data in these compressed formats, such approaches are mainly practical for highly sparse matrices, 99\% or greater.

In DNNs, significant overheads are introduced by the indirection required by the indexing. For each non-zero element in the sparse matrix, its location in memory must be determined by consulting the index before a corresponding element in the other matrix is loaded. In addition, the irregular nature of sparse matrices in DNNs removes many locality benefits (limiting the benefits from data caches and vector engines). Adjacent elements in the sparse weight matrix no longer necessarily interact with adjacent elements in the activation matrix, requiring serial processing or costly scatter-gather operations.

\subsubsection{Structured Sparsity}
\label{sec:structured_sparsity}
These problems with sparse matrices have been long recognized, and techniques have been developed to increase locality and amortize the overheads associated with the indices. These techniques constrain the locations at which non-zero elements can appear in the sparse matrices. There are two techniques commonly used to introduce structure and make sparsity more hardware friendly:
\begin{itemize}
\item \textbf{Partitioned Sparsity}: the matrices are divided into \textit{N} partitions, and conditions are placed upon the locations and number of non-zero elements in each partition~\citep{part,k5,k12}. 
\item \textbf{Block Sparsity}: Non-zero elements are forced to occur in blocks (typically square or rectangular). The indexing overhead is now amortized over the processing of all of the elements in the block, and a degree of locality is restored \citep{bulucc2009parallel}.
\end{itemize}

\begin{figure}[t]
    \centering
    \includegraphics[width=13cm]{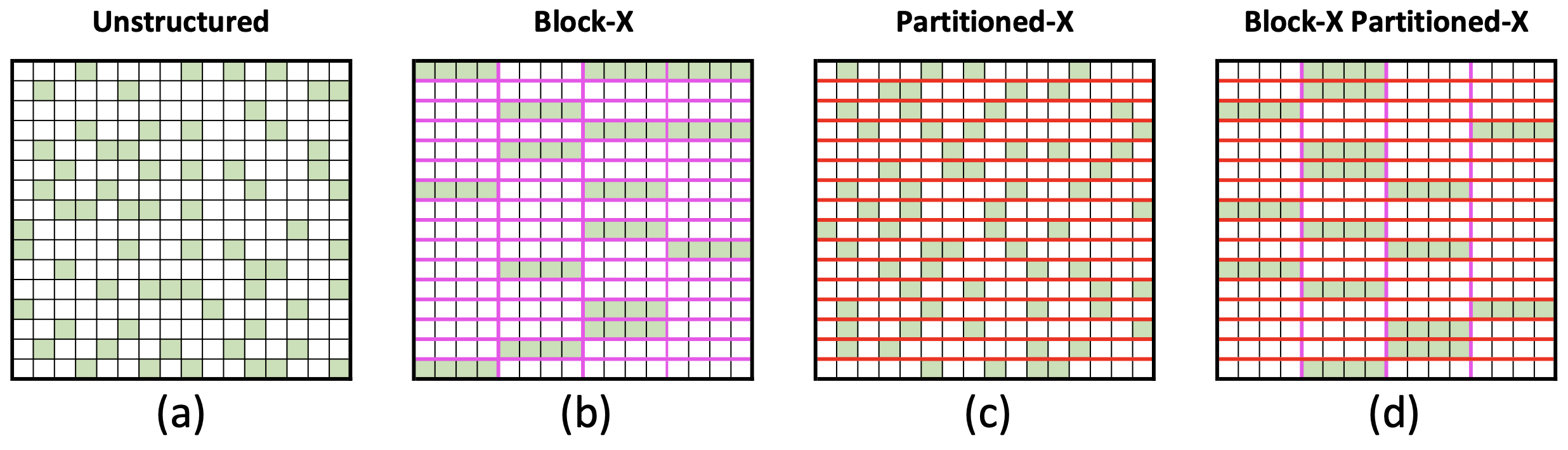}
    \caption{Examples of 75\% sparse matrices (a) illustrates unstructured sparsity in a matrix; non-zero elements are shaded, (b) illustrates block sparsity, with non-zero elements constrained to appear in 4-element blocks along a row, (c) illustrates partitioned sparsity, where each row is considered a partition and is forced to contain the same number of non-zero elements, while (d) applies both constraints.}
    \label{fig:block_sparse}
\end{figure}

\begin{figure}[t]
  \centering
  \subfloat[Sparse-Dense Matrix Computation]{\includegraphics[width=0.48\textwidth]{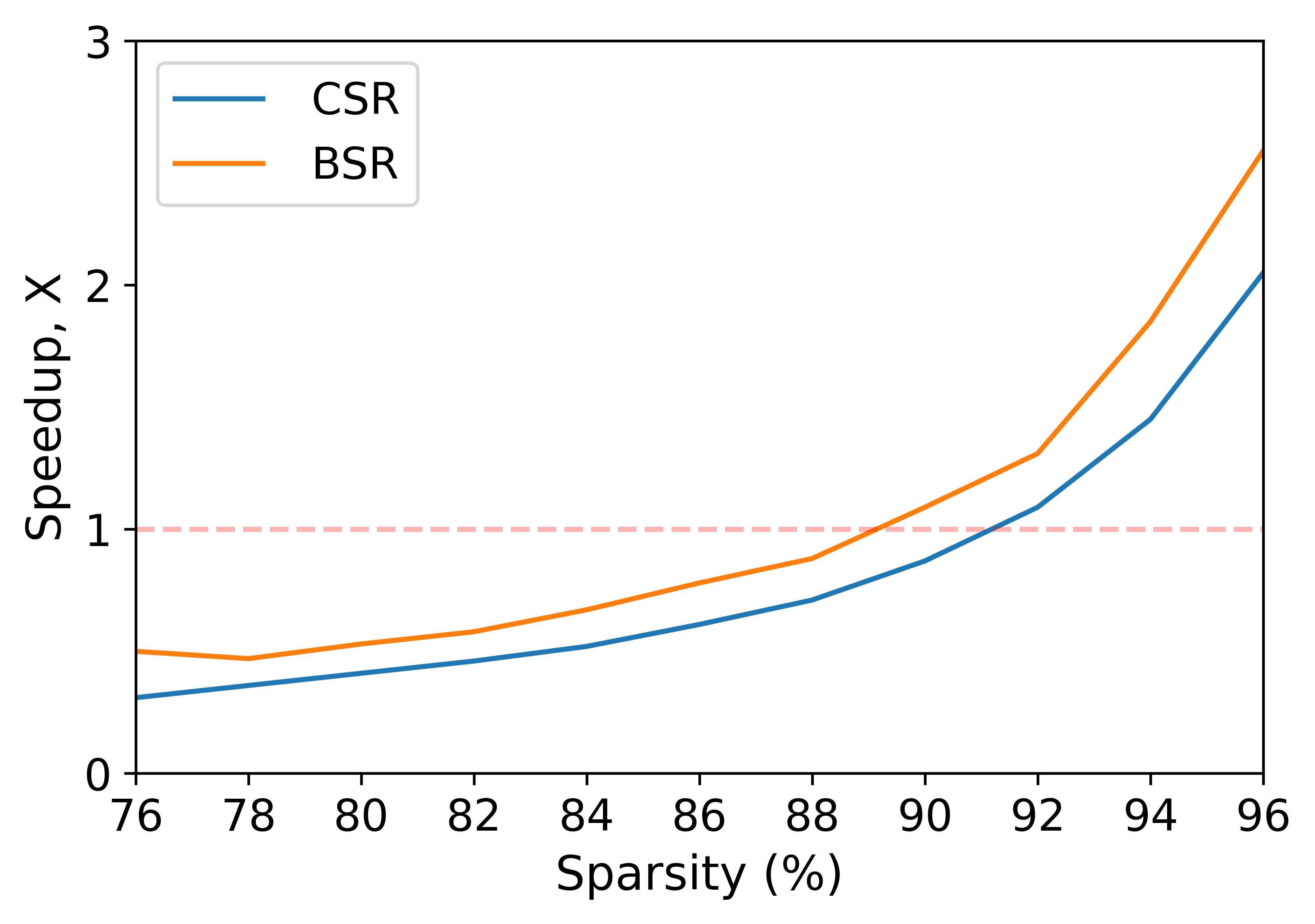}\label{fig:mkl1}}
  \subfloat[Sparse-Sparse Matrix Computation]{\includegraphics[width=0.48\textwidth]{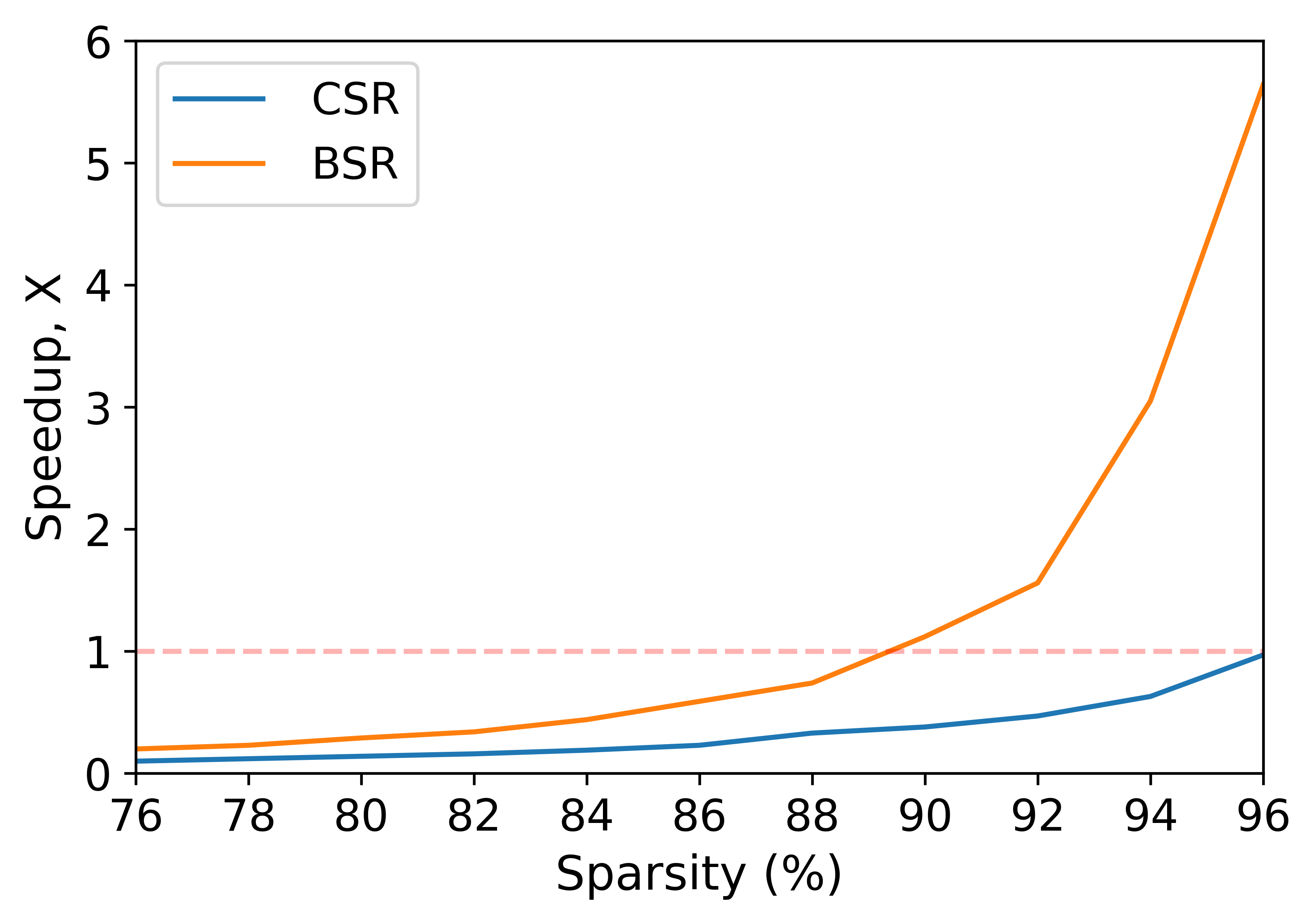}\label{fig:mkl2}}
  \caption{Performance improvements from OneAPI sparse matrix routines for both CSR and BSR (Block Compressed Row Format). The study was performed on an 8-core, AVX-512 capable processor (1024$\times$1024 matrix, with block size = 8$\times$8).}
\end{figure}

These techniques can be applied independently or in combination, with partitioning constraints imposed on the locations of the non-zero blocks, as illustrated in Figure~\ref{fig:block_sparse}. Imposing these constraints can provide significant performance benefits, as illustrated in Figure~\ref{fig:mkl1} and \ref{fig:mkl2}. In these figures the performance of sparse matrix multiplication (1024$\times$1024 matrix) operations is shown on a modern CPU processor using the highly tuned Intel OneAPI math libraries~\citep{oneapi}. It is apparent that imposing structure on the non-zero locations improves performance: for unstructured 96\% sparse matrices, sparse-dense multiplications are accelerated by around 2X, while no performance benefit is observed for spare-sparse multiplication. In contrast, when block sparse constraints are introduced, we achieve an almost 6X improvement for sparse-sparse operations. However, given that the sparse matrices are 96\% sparse, the theoretical computational savings associated with a sparse-dense operation is 25X, and 625X for the sparse-sparse operation, dwarfing the actual performance gains realized. It is also important to note that until a high degree of sparsity is obtained, matrix computations using the OneAPI sparse methods actually observed a slowdown, clearly demonstrating the challenges posed by sparse matrices on current hardware. Similar results are reported on GPUs using block sparse techniques~\cite{gpu-sparse}.

Imposing structure on the locations of the non-zero elements places limitations on the assignment of connected neurons and active neurons. There has been significant research into whether the imposition of these constraints degrades the achievable accuracy or peak sparsity. As might be expected, the impact is dictated by the severity of the constraints. If the partitions are too small, or the blocks too large, accuracy becomes degraded to an unacceptable extent~\citep{hug32}. 

Significant research has been invested in developing software optimization techniques to maximize the performance achievable with existing sparsity patterns \citep{Gray2017, batch}. However, while partitioned and block sparsity techniques have been partially successful in delivering the performance benefits of sparse DNNs on current hardware, these sparse techniques fall far short of their theoretical potential.

\subsubsection{Additional Challenges For Activation Sparsity}
Further to the costs of actually performing the sparse matrix multiplications, there are overheads associated with first determining which elements should be non-zero. During inference, the neuron weights are static, allowing the non-zero elements and resulting compressed representations to be computed offline. In contrast, for sparse activations, the non-zero elements are input dependent, and must be repeatedly recomputed during inference. Once the non-zero elements have been determined, there is an overhead for generating an appropriate representation of the sparse activations. For $k$-WTA, determining the winning elements involves performing a computationally costly sort operation, although partitioning or blocking schemes can be used to reduce the overheads. These overheads are not incurred when activations are dense, and represent a significant obstacle to achieving speedups from activation sparsity.

\section{Complementary Sparsity}
\label{sec:complementary_sparsity}
Directly processing a native representation of a sparse matrix is inefficient because of the presence of the zero-valued elements. Block and partitioned sparsity help align the patterns of non-zero elements with hardware requirements, but are fundamentally at odds with creating highly sparse and accurate networks. Optimal performance requires large blocks and reduced partition sizes but this limits both the obtainable sparsity and the accuracy~\citep{hug32}. This in turn compromises the approaches from achieving the theoretical performance benefits of highly sparse networks.

We propose an alternate approach that inverts the sparsity problem by structuring sparse matrices such that they are almost indistinguishable from dense matrices. We achieve this by overlaying multiple sparse matrices to form a single dense structure. An optimal packing can be readily achieved if no two sparse matrices contain a non-zero element at precisely the same location. Given incoming activations, we can perform an element-wise product with the incoming activations (a dense operation) and then recreate each individual sum.

We term this technique \emph{Complementary Sparsity}. Complementary Sparsity introduces constraints upon the locations of non-zero elements but it does not dictate the relative positions of the non-zero elements, nor does it dictate the permissible sparsity levels. The technique can be applied to convolutional kernels by overlaying multiple 2D sparse matrices from a layer's 3D sparse weight tensor. Importantly, the technique provides a path to linear performance improvements as the number of non-zero elements decreases, even for very high levels of sparsity. 

Figure \ref{fig:CK_conv}(a) illustrates the use of Complementary Sparsity for convolutional kernels. In this example, each kernel is 80\% sparse, and a set of 5 kernels with non-overlapping patterns is overlaid to form a single dense kernel. The number of sparse kernels that can be combined is directly proportional to their sparsity. The primary constraint is that the non-zero elements in each set should not collide with each other. Note that it is not necessary that all the weights in a layer are non-overlapping - the restriction applies only to each set being combined. Using our 80\% example, if a convolutional layer contains 20 channels, there are 4 dense sets each containing 5 sparse kernels. The elements must be complementary within a set, but there are no restrictions across the 4 sets. In addition, sets of convolutional kernels can be overlaid in either the filter or channel dimensions. Given this flexibility, in practice we have found that networks trained with the restrictions imposed by Complementary Sparsity do not compromise on accuracy when compared with unstructured sparsity.

Another important advantage of Complementary Sparsity is that it provides a path to facilitate both sparse weights and sparse activations. In the following subsections, we first describe the architecture of sparse-dense networks (i.e. networks with sparse weights and dense activations) and then describe the extension to sparse-sparse networks. Finally, we describe how these concepts can be implemented in an FPGA.

\subsection{Complementary Sparsity and Sparse-Dense Networks}

\begin{figure}
    \centering
    \includegraphics[width=13cm]{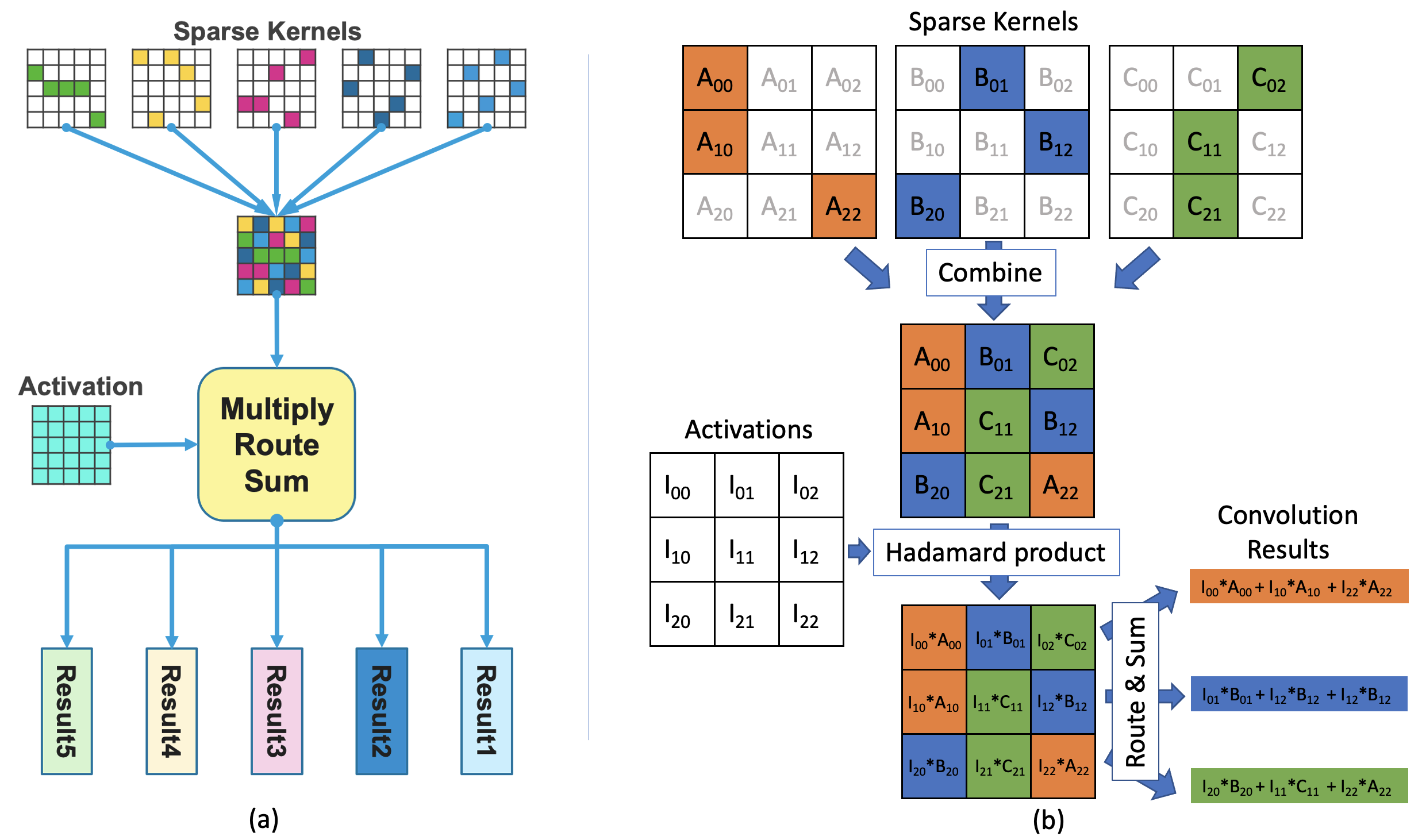}
    \caption{(a) 80\% Complementary Sparsity packs 5 sparse convolutional kernels into a single kernel for processing, (b) Complementary Sparsity applied in the filter dimension for 3x3 convolutional kernels. With 66\% sparsity, 3 sparse kernels are combined into a single dense kernel for processing.}
    \label{fig:CK_conv}
\end{figure}

The basic technique described above combines multiple sparse weight structures into a single dense entity, and natively supports sparse-dense networks, i.e. networks with dense activations and sparse weights. Partial results from each sparse entity must be kept separate and independently accumulated for final results. In sparse-dense networks processing is comprised of four distinct steps (Figure~\ref{fig:CK_conv}b):
\begin{enumerate}
    \item \textbf{Combine}: multiple sparse weight structures are overlaid to form a single dense entity. This is done offline as a preprocessing step.
    \item \textbf{Multiply}: each element of the activation is multiplied by the corresponding weight elements in the dense entity (Hadamard product).
    \item \textbf{Route}: the appropriate element-wise products are routed separately for each output.
    \item \textbf{Sum}: routed products are aggregated and summed to form a separate result for each sparse entity.
\end{enumerate}
The optimal techniques for implementing each component are dictated by the specifics of the target hardware. For example, in some cases, instead of routing the element-wise products, it may prove preferential to reorder the incoming activations. 

Given that Complementary Sparsity reduces $N$ sparse convolutions into a single dense operation, there is the potential for a linear $N$-fold performance improvement. The key challenge is to reduce the cost associated with routing and accumulating the packed results. Accordingly, of particular interest are techniques focused on minimizing the overheads associated with the routing of the Hadamard sub-products. Implementation of arbitrary routing usually involves resource hungry crossbar modules, where footprint increases as the square of the number of inputs. However, for DNN inference operations, the locations of the non-zero elements have been determined during training, and remain static throughout inference. The required routing is both fixed and predetermined, ensuring efficiency by tailoring implementations to the specific requirements of the network. 

To further minimize the overheads associated with the routing of the sub-products, Complementary Sparsity can be combined with the other forms of structural sparsity. For example, in Figure~\ref{fig:CK_conv}a, each column in the kernel is a partition, with one non-zero element permitted per column. Similarly, complementary patterns with blocks of non-zero elements are possible, as shown in Figure~\ref{fig:block_sparse}. Section~\ref{sec:sparse-routing} below describes our FPGA implementation of routing in more detail, and Section~\ref{sec:resnet_results} analyzes resource tradeoffs.

\subsection{Complementary Sparsity and Sparse-Sparse Networks}
\label{sec:sparse-sparse}
The above sparse-dense Complementary Sparsity technique can be extended to handle sparse-sparse networks, i.e. networks comprised of both sparse activations and sparse weights. As discussed in Section~\ref{sec:hw_bad}, significant inefficiencies are traditionally associated with sparse-sparse matrix computations due to the changing locations of non-zero elements in the activation vectors. For optimal performance, these non-zero activations must be paired with the respective non-zero weights. The overheads associated with dynamically interpreting the sparse representations, and then ensuring they rendezvous with the appropriate sparse non-zero weights severely degrades any performance gains associated with processing the sparse subset of elements. 

\begin{figure}
  \centering
  \subfloat[]{\includegraphics[width=0.5\textwidth]{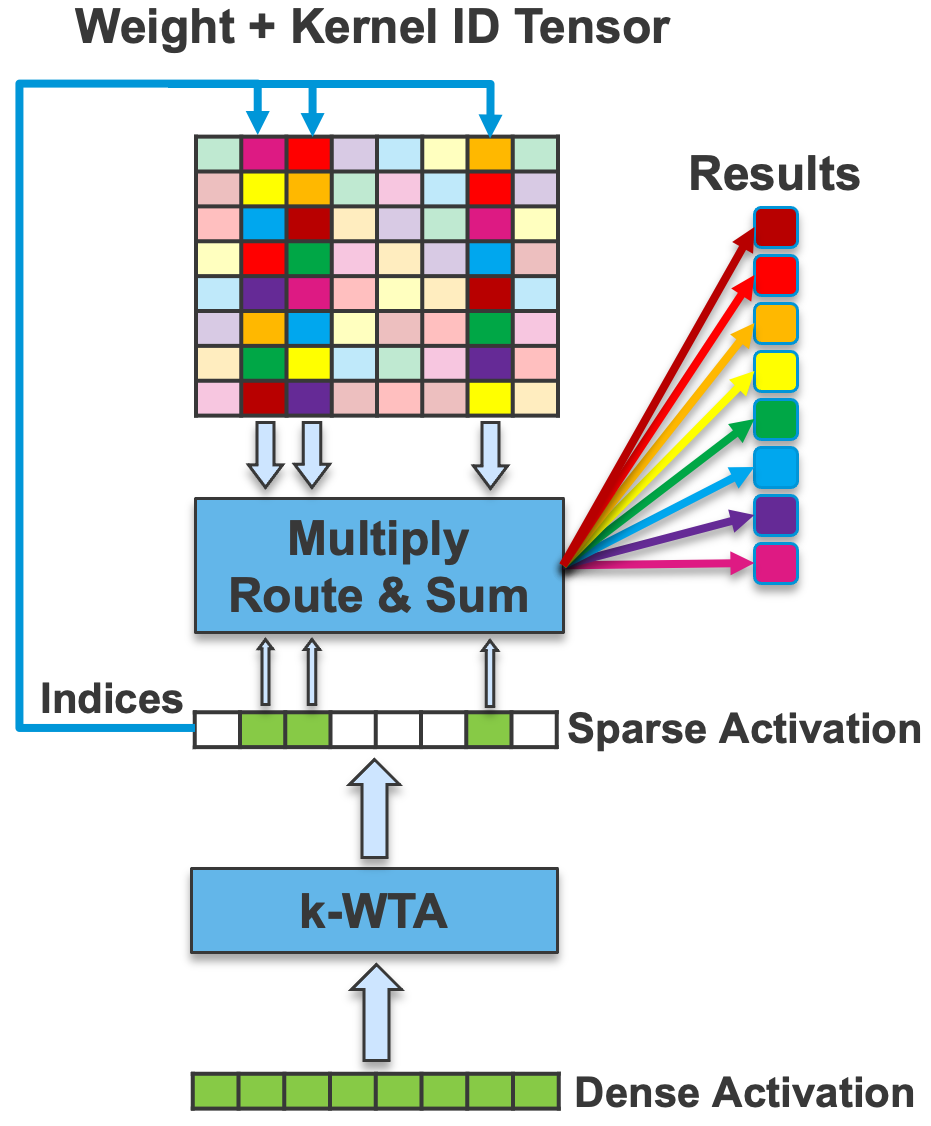}\label{fig:spsp}}
  \subfloat[]{\includegraphics[width=0.4\textwidth]{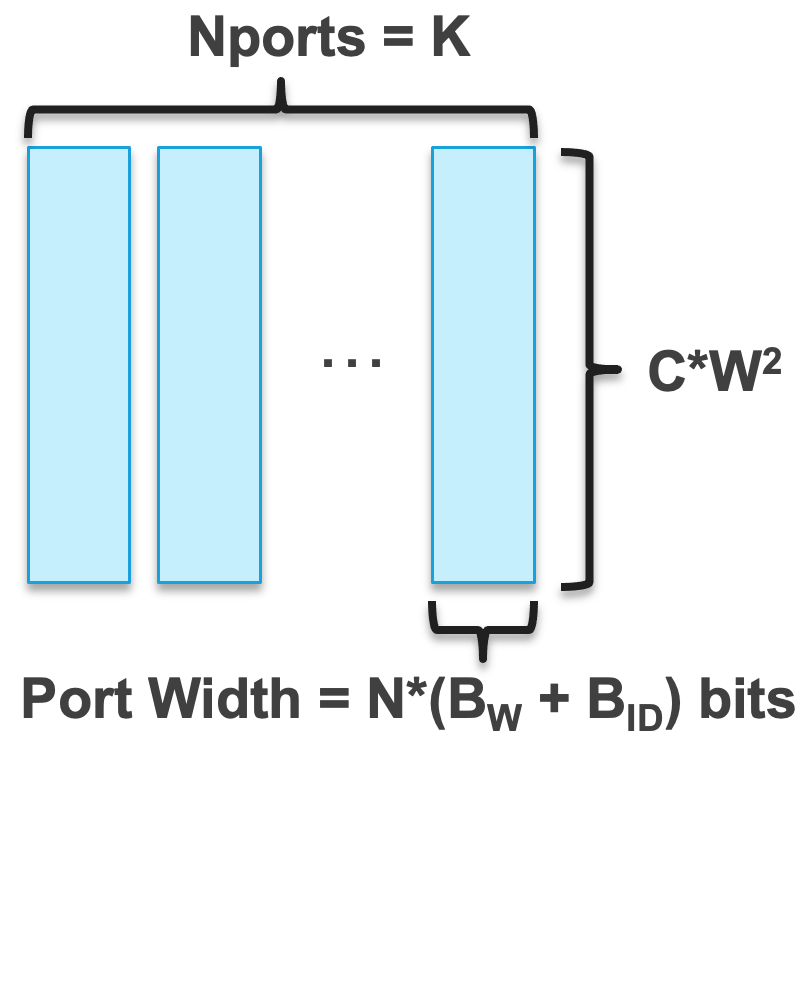}\label{fig:wtd}}
  \caption {(a) With Complementary Sparsity, sparse weight tensors are represented as simple dense tensors, simplifying the addition of activation sparsity, (b) An illustration of the Sparse-Sparse weight tensor structure, C is the number of channels, W is the width of the kernel, N is the number of non-zero weights per kernel, B\textsubscript{W} is the bit width of the weight values, K is the number of activations processed in parallel, and B\textsubscript{ID} is the bit width of the Kernel ID.}
\end{figure}

Using Complementary Sparsity, the sparse-sparse problem is simplified to a problem with sparse activations and dense weights. Since the weights are dense, the above overheads are eliminated. As illustrated in Figure \ref{fig:spsp}, when the sparse weights are represented in a dense format, the incoming sparse activations are paired with the relevant weights. For each non-zero activation there exists a corresponding column of non-zero weight elements at a predefined location in the dense weight structure. Processing is comprised of the following five steps:

\begin{enumerate}
    \item \textbf{Combine}: multiple sparse weight structures are overlaid to form a single dense structure. This is done offline as a preprocessing step.
    \item \textbf{Select}: a $k$-WTA component is used to determine the top-$k$ activations and their indices. 
    \item \textbf{Multiply}: each non-zero activation is multiplied by the corresponding weight elements in the dense structure.
    \item \textbf{Route}: the appropriate element-wise products are routed separately for each output.
    \item \textbf{Sum}: routed products are aggregated and summed to form a separate result for each sparse matrix.
\end{enumerate}

Compared to sparse-dense, the extensions are in the second and third steps. This formulation directly takes advantage of sparse activations. The computation in the third step is reduced in proportion to the sparsity of the incoming activations. As with the sparse-dense case, a key issue is the additional overhead imposed by routing. For sparse-sparse there is also additional overhead imposed by the $k$-WTA block. An efficient implementation of these components is critical to realizing an overall benefit, and are detailed below in Sections~\ref{sec:sparse-routing} and \ref{sec:fwta}.

\subsection{Complementary Sparsity on FPGAs}
\label{sec:cs_FPGA}
In this section, we discuss our implementation of Complementary Sparsity on FPGAs, before presenting both performance and resource utilization results in Sections~\ref{sec:gsc_results} and~\ref{sec:resnet_results}. We focus our discussion on the sparse-sparse implementation of convolutional kernels, specifically the individual components of Figure~\ref{fig:spsp}. The implementations are focused on inference operations, and assume the use of fixed-point arithmetic. As discussed, the flexible architecture of FPGAs represents a model platform for exploring idealized circuit structures for Complementary Sparsity.

\subsubsection{Sparse-Sparse Hadamard Product Computation}
\label{sec:hada}

For each non-zero activation the relevant weights must be extracted from the weight tensor, and then individually multiplied by the activation (Hadamard product, third step in Section~\ref{sec:sparse-sparse}), before being passed to the next stage for routing. The sparse filter kernels, once converted into a set of dense kernels, are retained in a series of memories on the FPGA. As illustrated in Figure~\ref{fig:spsp}, the indices associated with the non-zero elements of an incoming sparse activation are used to directly extract the relevant elements from the weight tensor. For each non-zero activation value, we chose to retrieve all relevant weights in the weight tensor in parallel. Associated with each weight is a Kernel ID (identifying the channel index in the output tensor) that is subsequently used to route the resulting sub-product. The Kernel ID index is determined \textit{a priori}, and is attached to the corresponding weight value, forming an augmented tensor.
 
Processing the incoming activation tensor in parallel delivers the greatest convolution throughput. However, this parallelism comes at a cost, requiring parallel accesses to the augmented weight tensor. Assuming $K$ non-zero activation values, and, given that the indices of each value are dynamic, a $K$-ported weight tensor memory is required to retrieve each set of weights in parallel, as illustrated in Figure~\ref{fig:wtd}. The figure demonstrates that each port must also be sufficiently wide to handle the multiplicity of kernels associated with an activation index. 

Note that, as activation sparsity is increased, the number of ports required decreases. Similarly, the higher the weight sparsity, the smaller the port size for each of those memories. Both requirements are approximately proportional to the degree of sparsity. In Figure~\ref{fig:wtd}, the factor $N$ associated with the port width is the number of dense complementary filter weight vectors that must be accessed in parallel. 

\begin{figure}[t]
  \subfloat[]{\includegraphics[width=0.5\textwidth]{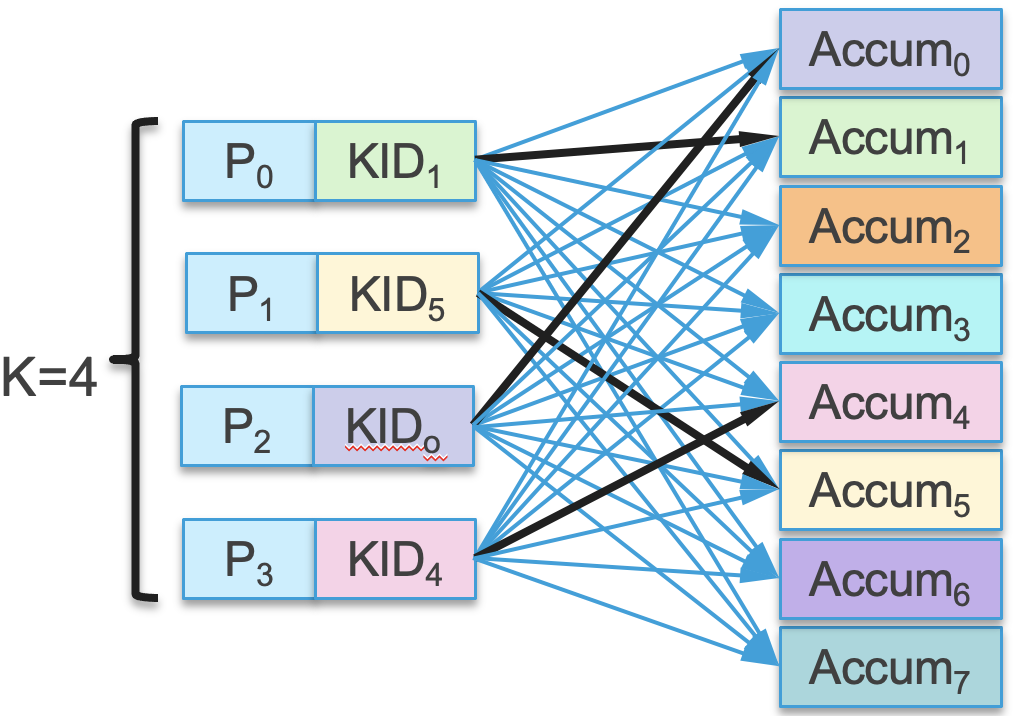}\label{fig:praa}}
  \subfloat[]{\includegraphics[width=0.5\textwidth]{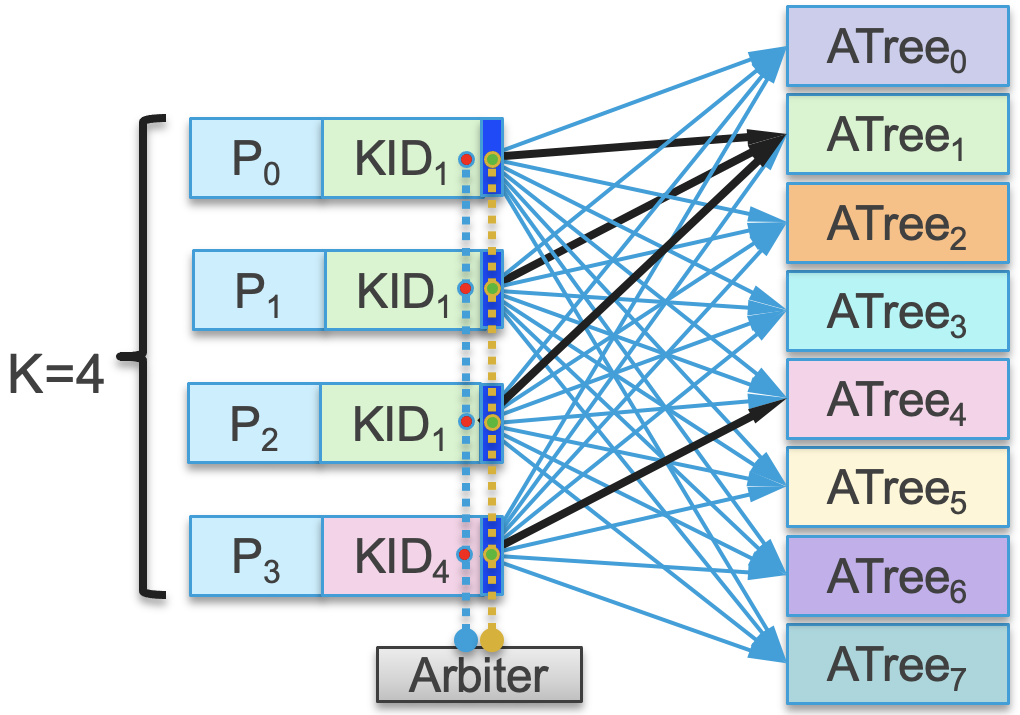}\label{fig:wta-adder}}
  \caption {Sparse-Sparse sub-product routing: (a) Sub-products serially routed to the appropriate accumulator, (b) Sub-products routed in parallel to the appropriate adder-tree.}
\end{figure}

\subsubsection{Sparse-Sparse Hadamard Product Routing}
\label{sec:sparse-routing}
A second critical component is the efficient routing of the sub-products from the Hadamard operation. Once an activation has been multiplied by the retrieved weights, to complete the computation of the convolution, each resulting sub-product must be combined with the other sub-products from the same kernel. The relevant products are identified using their Kernel ID tag.

For $K$ non-zero activation vectors, the retrieved weights may belong to a single filter kernel (identical Kernel IDs), or might be distributed across several filter kernels. Each of the $K$ activations can be processed serially, in which case the results for each of the products can be simply routed via a multiplexor network to an accumulator, based upon its Kernel ID. This is diagrammed in Figure~\ref{fig:praa}.

For greater performance, the products from all the activations can be processed in parallel. In this case, the resulting sub-products must be routed to an adder tree for summing, rather than to a single accumulator (Figure~\ref{fig:wta-adder}). Routing of multiple addends to non-conflicting locations in an adder tree introduces additional complexity. Not only is it necessary to route based upon the Kernel ID, but additional destination address bits are required to designate the slot of the adder in which the product should land. This is resolved with an arbitration module, which supplies these additional address bits before the product is passed to a larger multiplexor network. The arbitration module generates the low order address bits from the set of Kernel IDs, using a prefix sum algorithm. Each occurrence of a product with the same Kernel ID increments the value of the lower order bits, ensuring they are assigned to a non-conflicting slot in the adder tree for that kernel. Partitioning reduces the size of these indices since we only need sufficient bits to identify their position within the partition, not within the entire activation or output tensor.

\subsubsection{Activation sparsity using \textit{k}-WTA}
\label{sec:fwta}
For $k$-WTA, activation sparsity is induced by explicitly restricting the number of non-zero elements to the $K$ largest values produced by a layer. Determining these top $K$ values efficiently can represent a significant obstacle to the effective use of activation sparsity. The time and resources expended performing the sort operation erodes the performance benefits associated with leveraging the resulting sparsity in subsequent processing. Accordingly, an optimized $k$-WTA implementation is central to our FPGA implementation.
\begin{figure}
    \centering
    \includegraphics[width=1.0\textwidth]{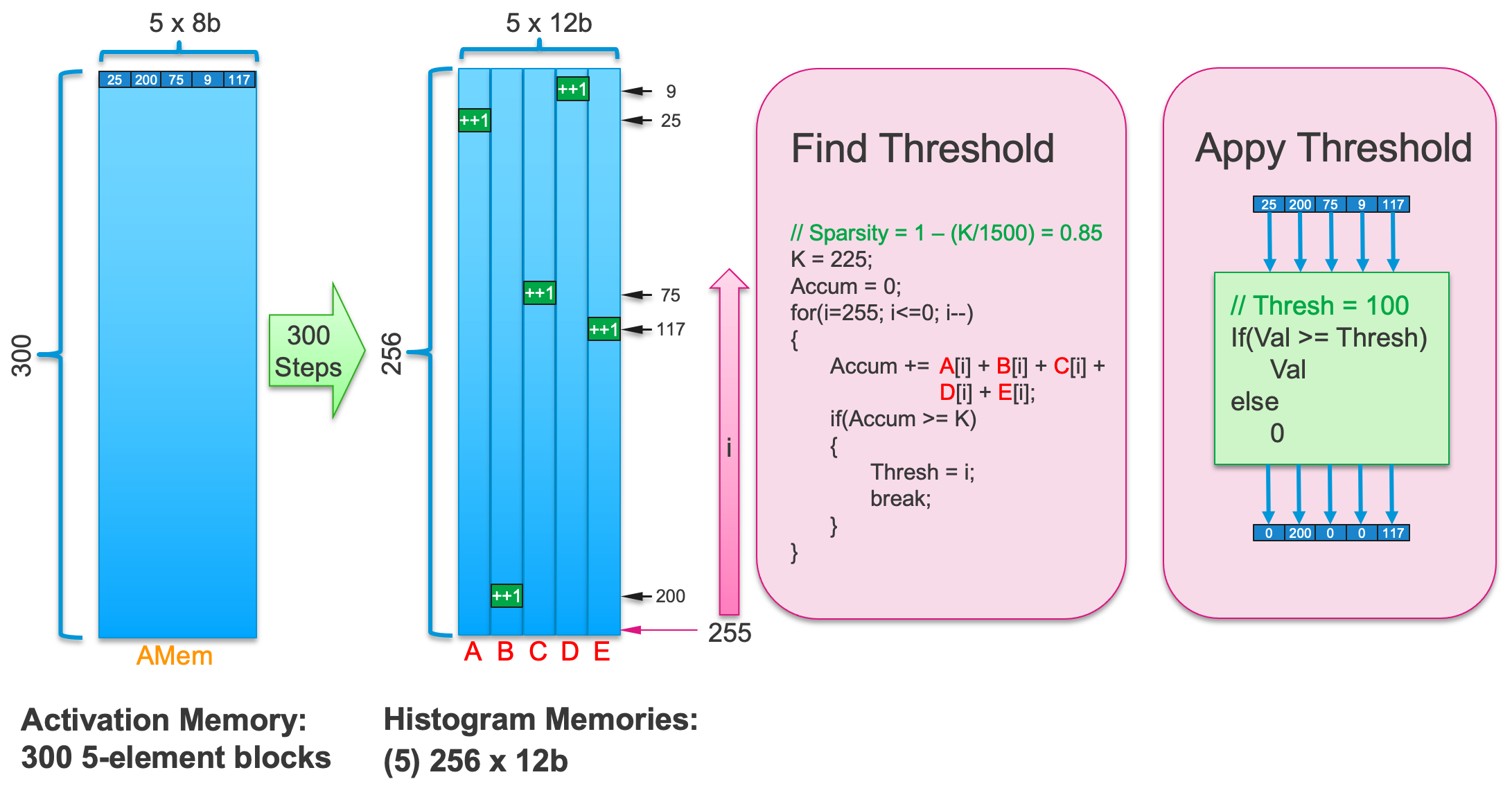}
    \caption{Parallel Global $k$-WTA: performs a histogram based search of entire activation tp determine the threshold yielding K non-zero elements. In this example, the 1500-element activation is stored as 300 5-element blocks in memory AMem. Each block is read out and the element values are used to address and increment counts in 5 separate memories, A-E. The counts are then cumulatively summed together into variable Accum, starting with the largest value location, until Accum reaches a total count of K, establishing a thresholding value. Values in AMem are then compared to Thresh, and sent through to the next layer if they are greater than or equal to Thresh, along with their indices in the original 1500 element vector.}
    \label{fig:bgkwta}
\end{figure}
We divide $k$-WTA implementations into two broad categories:
\begin{itemize}
    \item \textbf{Global}: All elements of an activation are examined to determine the $K$ largest. We use global $k$-WTA following linear layers.
    \item \textbf{Local}: The activation is partitioned into smaller units, and only the elements belonging to a partition are compared to each other. We use local $k$-WTA following convolutional layers, where the winner take all competition happens along the channel dimension.
\end{itemize}

For 8-bit activation values, our implementation of global $k$-WTA leverages a histogram-based approach. In our implementation, a 256-element array in memory is used to build the histogram, with each activation value being used to increment a count at a location addressed by that value. Once all of the activation values have been processed, the histogram array represents the distribution of the activation values. For a specified value of $K$, the histogram values can be read, largest first, to determine the appropriate minimum value cutoff; values above this threshold should be retained as part of the top-$k$ and the remainder discarded. As a final step, the activation values are compared against the threshold and the winners passed to the next layer. 

For improved performance, an implementation may process multiple activation elements in parallel. In this scenario, multiple histograms are built in parallel and then combined to determine the overall cutoff value. An example of this implementation is illustrated in Figure~\ref{fig:bgkwta}, for 1500-activations, 5-way parallelism, and activation sparsity of 85\%.

\begin{figure}
    \centering
    \includegraphics[width=1.0\textwidth]{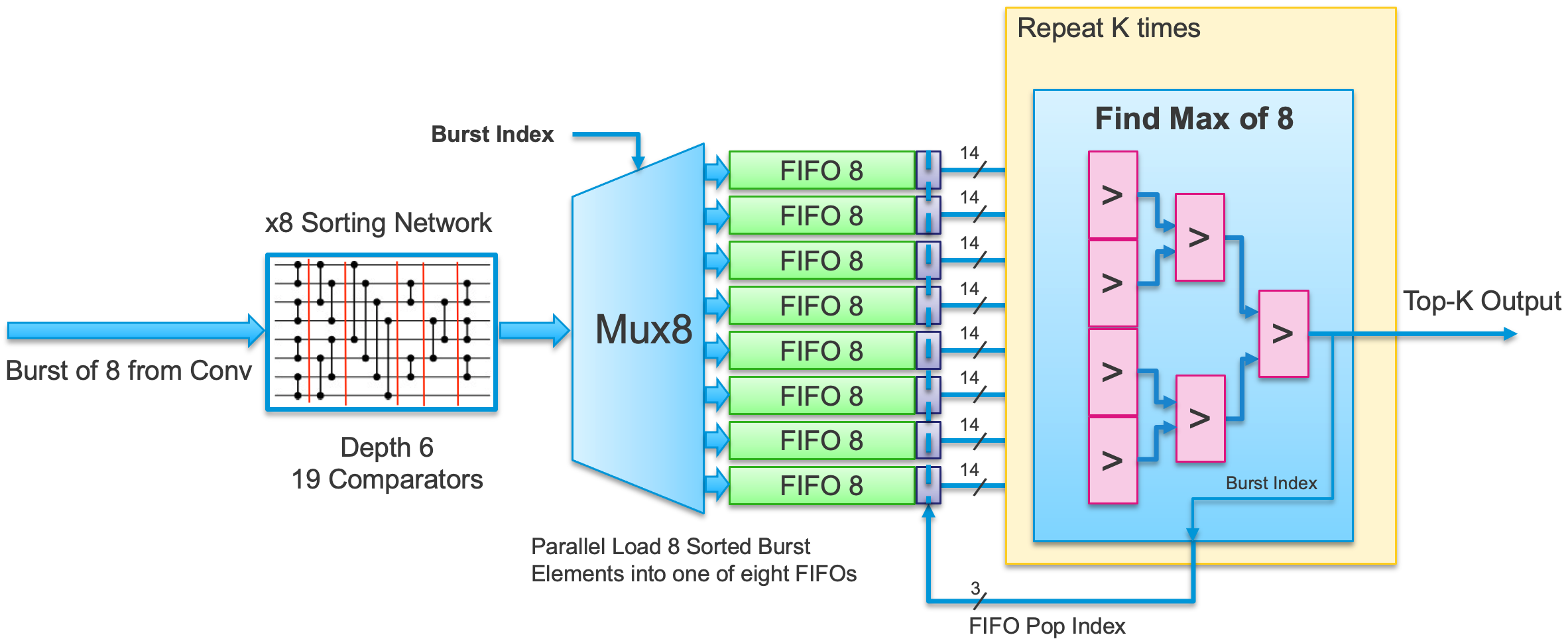}
    \caption{Structure of $k$-WTA when serially processing Complementary Sparse Convolutions.}
    \label{fig:bkwta}
\end{figure}

For convolutional layers, activations have a natural partitioning in the channel dimension. When the top-$k$ operation in $k$-WTA is implemented as a sorting operation, which is $O(N*log(N))$ either in time or hardware resources, partitioning provides significant efficiency benefits. The position of each result value produced by the convolutional layer must be tracked through the sorting process. This is achieved by appending an index to each data value entering the sorting function. 

Sorting is performed in several stages, and the implementation is optimized based on the key observation that it is only necessary to find the top $K$ values in each vector. The ordering of the low valued elements is immaterial, and, as $K$ decreases with increasing activation sparsity, we ensure that the cost of sorting implementation falls accordingly. First, each vector is subdivided into $M$ sub-vectors, and each sub-vector is sent through a sorting network. The sorted sub-vector is subsequently loaded into one of $M$ FIFOs, with each sub-vector’s largest value at the front of its FIFO.

A vector composed from the $M$ top-of-FIFO values is then passed through a $log_2(M)$ stage comparator tree, in order to determine the maximum value in the vector. The maximum value is retained, and its associated indexing information (which indicates in which FIFO the value was located) is used to pop that element from the appropriate FIFO, exposing the FIFO's next largest element. This process is repeated $K$ times; at which point the output vector has been filled with the top $K$ elements and is passed to the next processing layer. 
In our implementation, a 64-element vector is subdivided into eight 8-element sub-vectors. The sorting network is comprised of 19 comparators, arranged into depth 6 layers. There are 8 FIFOs, and a 3-level comparator tree is used to determine the maximum value in the 8-element top-of-FIFO vector.

\begin{figure}
    \centering
    \includegraphics[width=1.0\textwidth]{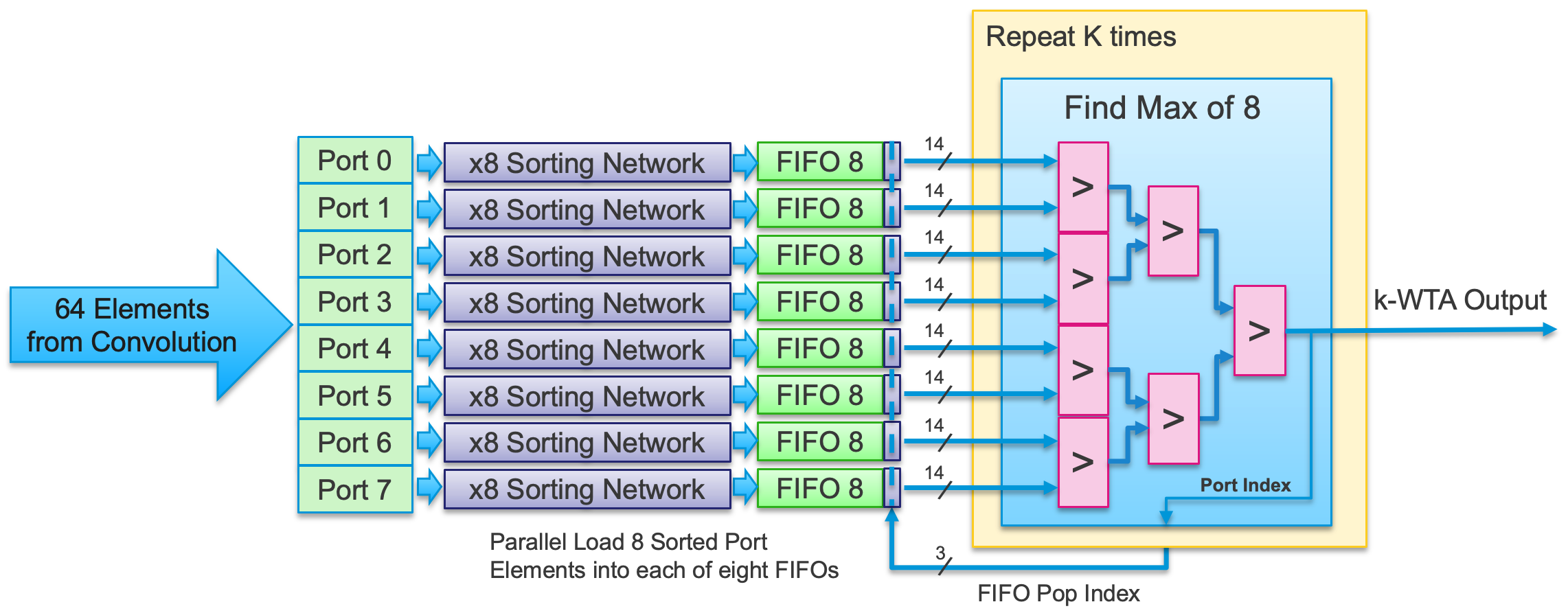}
    \caption{Structure of $k$-WTA when parallel processing Complementary Sparse Convolutions.}
    \label{fig:pkwta}
\end{figure}

To prevent bottlenecks, the performance of the $k$-WTA implementation should be matched to the performance of the convolutional operator. The incoming results can either arrive in serial bursts or as complete result vectors. A $k$-WTA implementation could wait until all bursts have been concatenated and a complete activation result is available, or take advantage of the burst intervals and combinationally sort~\cite{knuth} the burst values before loading it into one of the FIFOs. The serial burst implementation is illustrated in Figure~\ref{fig:bkwta}. Alternatively, all the activation results could be computed in parallel, partitioned into $M$ groups, and pushed through $M$ instances of a combinational sort before being loaded in parallel to the $M$ FIFOs, as illustrated in Figure~\ref{fig:pkwta}.

In summary, there are a number of ways to implement $k$-WTA efficiently. By choosing appropriately we find that overall the $k$-WTA is a relatively small percentage of overall resource usage (see Section~\ref{sec:kwta-resou}).

\section{Results on an End to End Speech Network}
\label{sec:gsc_results}
\label{sec:gsc-desc}
In this section we discuss the application of Complementary Sparsity to an end to end speech recognition system. We trained a convolutional network~\citep{Rawat2017} to recognize one-word speech commands using the Google Speech Commands (GSC) dataset~\citep{gsc}. We implemented dense and sparse versions of the network on both large and small FPGA platforms. Our goal was to study the impact of Complementary Sparsity on full system throughput (the number of words processed per second) and understand trade-offs in resources, memory consumption and energy usage.

GSC consists of 65,000 one-second long utterances of keywords spoken by thousands of individuals. The task, to recognize the spoken word from the audio signal, is designed for embedded smart home applications that respond to speech commands. State of the art convolutional networks on this dataset achieve accuracies (before quantization) of 96-97\% using $10$ categories \citep{sainath2015convolutional,Tang2017}. 

Our base dense GSC network is a standard convolutional network composed of two convolutional layers, a linear hidden layer plus an output layer, as described in Table \ref{tab:GSC}. We also trained a sparse network with identical layer sizes but with both sparse weights and sparse activations. Our sparse network follows the structure and training described in \citep{Ahmad2019}. To enforce sparse weights we used a static binary mask that dictates the locations of the non-zero elements and meets requirements of Complementary Sparsity. The ReLU activation function was replaced by a ($k$-WTA) \citep{Makhzani2013, Ahmad2019} activation function (see Section~\ref{sec:activation_sparsity} and Figure~\ref{fig:sparsenet}). 

The baseline dense version of the network contained 2,522,128 parameters, while the sparse network contained 127,696 non-zero weights, or about 95\% sparse. The activations in the sparse network range from 88\% to 90\% sparsity (i.e. 10-12\% of the neurons are ‘winners’), depending on the layer. Both dense and sparse models were trained on the GSC data set, achieving comparable accuracies (see \citep{Ahmad2019} for details). In our implementation, the accuracies of the sparse and dense networks are between 96.4\% and 96.9\%. Both activations and weights are quantized to 8-bits.

\begin{table}[b]
\caption{\label{tab:GSC}Architecture of the CNN network trained on GSC data}
\begin{center}
\begin{small}
    \begin{tabular}{lcccc}
        \toprule
        \textbf{Layer} & \textbf{Channels} & \textbf{Kernel Size} & \textbf{Stride} & \textbf{Output Shape} \\
        \midrule
         Input & - & - & - & 32$\times$32$\times$1 \\
         Conv-1 & 64 & 5$\times$5$\times$1 & 1 & 28$\times$28$\times$64 \\
         MaxPool-1 & - & 2$\times$2$\times$1 & 2 & 14$\times$14$\times$64 \\
         Conv-2 & 64 & 5$\times$5$\times$64 & 1 & 10$\times$10$\times$64 \\
         MaxPool-2 & - & 2$\times$2$\times$1 & 2 & 5$\times$5$\times$64 \\
         Flatten & - & - & - & 1600$\times$1 \\
         Linear-1 & 1500 & 1600$\times$1 & - & 1500$\times$1 \\
         Output & 12 & 1500$\times$1 & - & 12$\times$1 \\
         \bottomrule
    \end{tabular}
\end{small}
\end{center}
\end{table}

\subsection{FPGA Implementation}
We implemented the baseline dense GSC network using the Xilinx\textsuperscript{TM} software “Vitis AI”~\cite{vitis_ai}. Vitis AI is the preferred solution for deploying deep learning networks on Xilinx FPGA platforms. Convolution and linear layers in Vitis AI invoke hand-optimized Processing Elements (PE) implemented using RTL. A software compiler converts a given network, including parameters and weights, into schedules of calls to these PEs.

We implemented our sparse GSC networks using the Xilinx Vivado HLS toolset~\cite{pragma, vivado}. Although HLS uses a C++ compiler (with Xilinx specific pragmas) and does not produce hand-optimized designs, it represents a faster design path. There was sufficient flexibility in the toolset to implement our sparse designs. We note however that the results for our sparse networks below would likely be improved using hand-optimized designs.

We created two pipelined implementations of our sparse network using HLS. The \emph{Sparse-Dense} implementation leveraged weight sparsity in Conv-2 and the linear layer, ignoring sparse activations. The \emph{Sparse-Sparse} implementation leveraged both sparse activations and sparse weights (as described in Section~\ref{sec:cs_FPGA}). In the Sparse-Dense implementation, the Conv-1 layer was left as fully dense as its profile was small relative to the other pipeline stages. In the Sparse-Sparse implementation the other stages became faster and Conv-1 became a bottleneck. As such we implemented Conv-1 using a sparse-dense strategy (the input to the network is dense, hence sparse-sparse is not an option for Conv-1).

\subsection{Benchmark Description}
The performance of the 3 different CNN implementations were tested on two different Xilinx FPGA platforms. The first, the Alveo U250~\cite{u250}, is a high-end card targeted at data centers, while the second, the UtraScale ZU3EG~\cite{zynq}, is a smaller system targeted at embedded applications. Compared to the ZU3EG, the U250 has 11X the number of system logic cells, about 56X the internal memory, and consumes 9X more power.

For each CNN network on each FPGA two different experiments were undertaken:
\begin{enumerate}
    \item \textbf{Single network performance}: a single network is a pipelined implementation of one GSC network, processing a single stream of speech commands. 
    \item \textbf{Full chip performance}: multiple network instances are placed on the FPGA until the entire FPGA's resources are exhausted, or the design cannot be routed by the software. Multiple input streams are distributed across the instances, and the inference throughput delivered by the entire chip is reported.
\end{enumerate}

In both experiments, the input data is a repeating sequence of 50,000 pre-processed audio samples. We chose overall throughput, measured as the total number of input words processed per second, as the primary performance metric.

\subsection{Single Network Results}
Table \ref{tab:single_network} shows the results of running a single network instance on the U250 and ZU3EG platforms. On the U250, the Sparse-Dense implementation achieves over 11.7X the throughput of the dense implementation, while the Sparse-Sparse implementation outperforms both the dense and the Sparse-Dense implementations by 33.6X and 2.8X respectively (Figures~\ref{fig:gsc1} and \ref{fig:gsc2}). 

The dense implementation did not fit on the smaller ZU3EG platform due to the limited resources available. Both sparse implementations were able to compile and run successfully on the platform due to their smaller size and lower resource requirements. The Sparse-Sparse implementation was about 2.1X faster than the Sparse-Dense implementation. Interestingly, the Sparse-Dense implementation on the ZU3EG platform was still 6.9X faster than the dense implementation on the more powerful U250. This demonstrates the performance benefits associated with sparse networks, and also the potential for sparse networks to open up new applications in embedded scenarios that were previously impossible.

\begin{table}[]
 \caption{Throughput of single sparse and dense networks on the U250 and ZU3EG platforms, measured in words processed per second. The dense network did not fit on the ZU3EG due to its limited resources. All sparse implementations, regardless of platform, were significantly faster than the dense network running on the U250. The Sparse-Sparse implementation was consistently 2X to 3X faster than the Sparse-Dense implementation.}
    \label{tab:single_network}
\begin{center}
\begin{tabular}{cccc}\toprule
\textbf{FPGA Platform} & \textbf{Network Implementation} & \textbf{Throughput} & \textbf{Speedup} \\\midrule
      & Dense & 3049 & 1.0  \\
     U250 & Sparse-Dense & 35,714 & 11.71 \\
      & Sparse-Sparse & 102,564 & 33.63 \\ 
      & & & \\
      & Dense & 0 & - \\
     ZU3EG & Sparse-Dense & 21,053 & N/A \\
      & Sparse-Sparse & 45,455 & N/A \\\bottomrule
\end{tabular}
\end{center}
\end{table}

\subsection{Full Chip Results}
\begin{figure}[!tbp]
  \centering
  \subfloat[]{\includegraphics[width=0.50\textwidth]{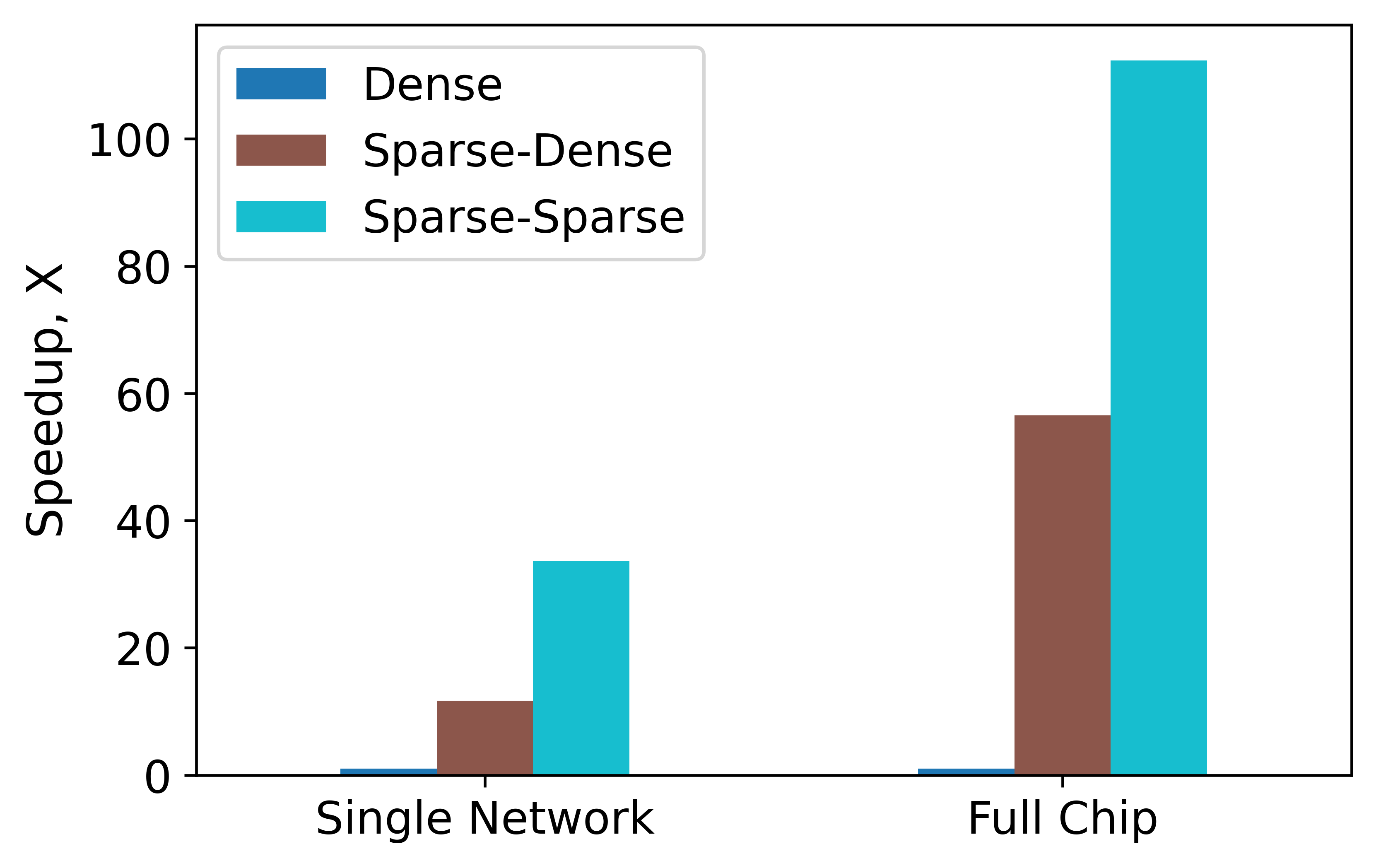}\label{fig:gsc1}}
  \subfloat[]{\includegraphics[width=0.50\textwidth]{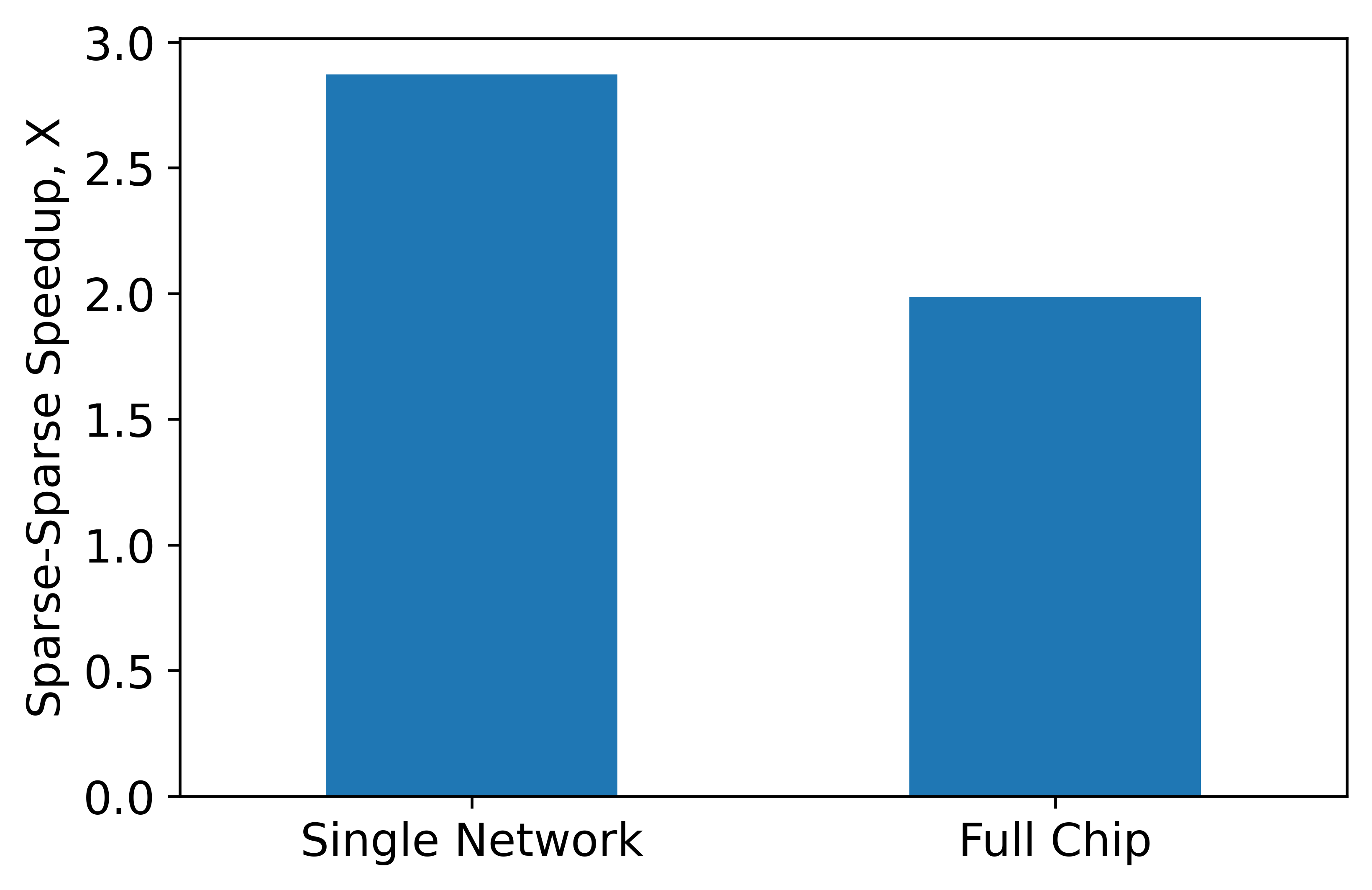}\label{fig:gsc2}}
  \\
    \subfloat[]{\includegraphics[width=0.50\textwidth]{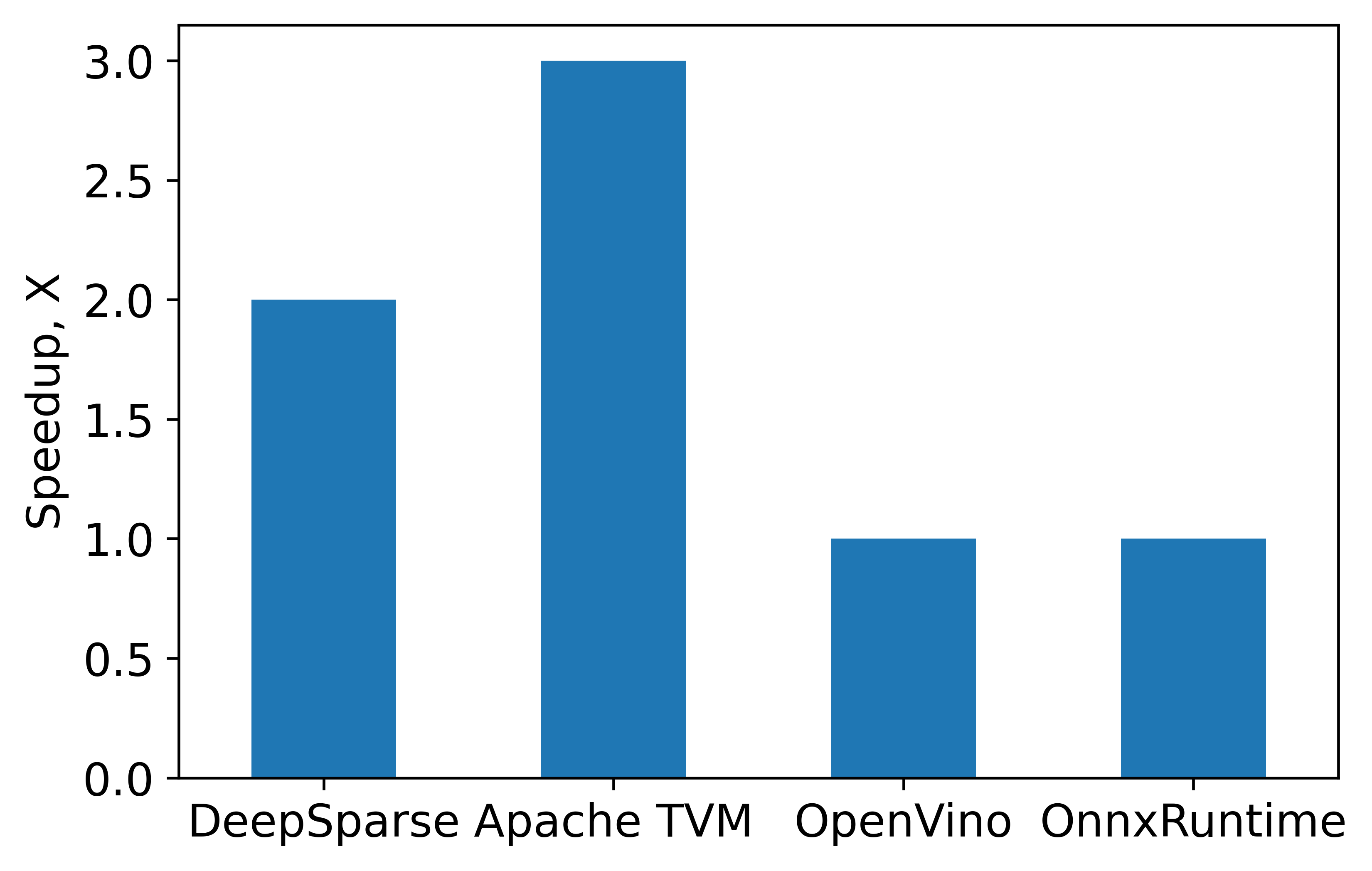}\label{fig:gsc3}}
  \subfloat[]{\includegraphics[width=0.50\textwidth]{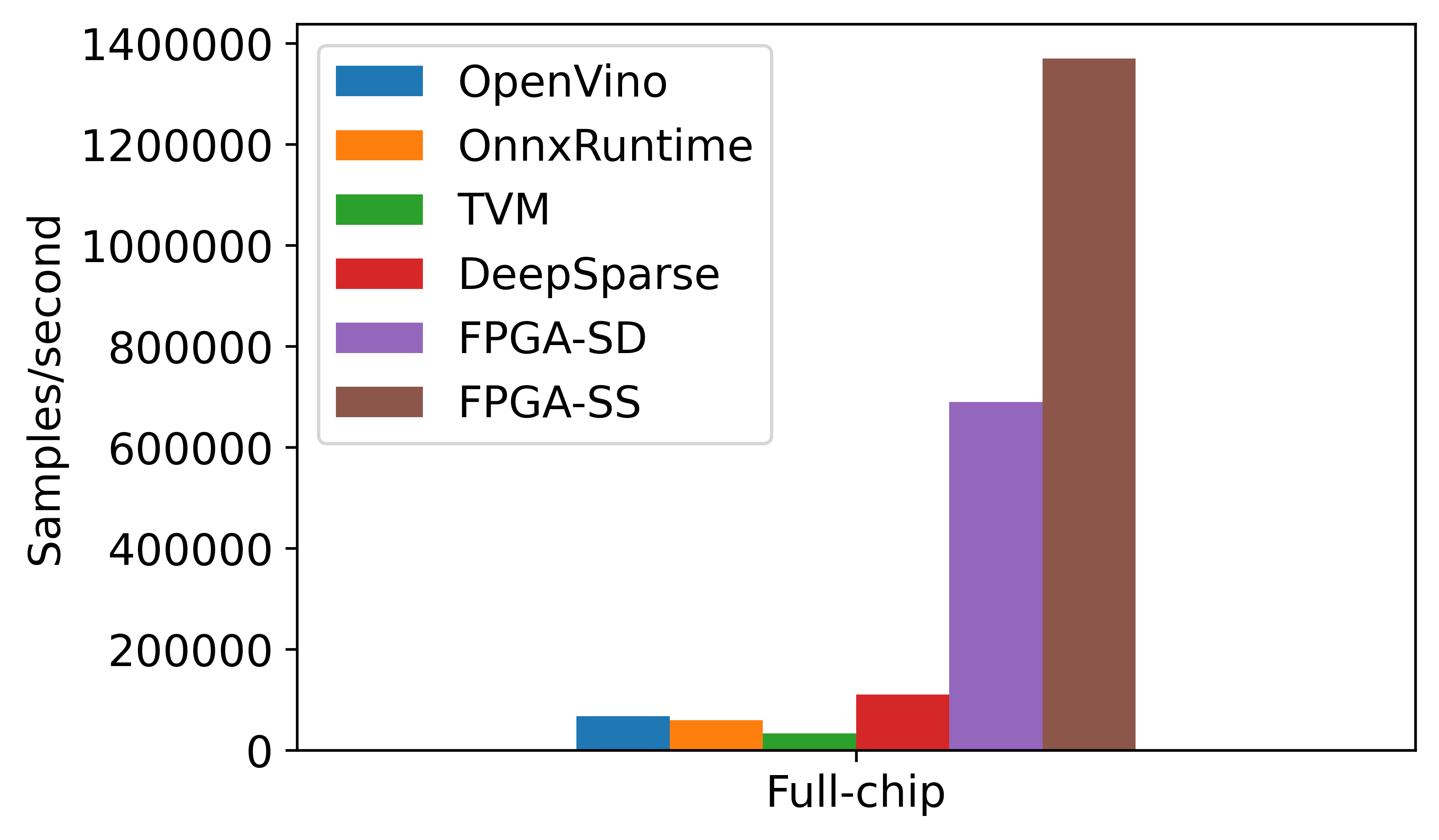}\label{fig:gsc4}}
  \caption{Performance comparisons between sparse and dense networks: (a) Sparse network performance on the U250 relative to dense, (b) Sparse-Sparse network performance, relative to Sparse-Dense, on the U250, (c) Sparse-Dense network performance on CPUs using common inference runtimes (relative to a dense network on the same runtime), (d) Sparse network performance on CPUs and FPGAs.}
\end{figure}

Table \ref{tab:full_chip} shows the full-chip throughput results for the U250. The numbers illustrate the performance benefits of sparse networks. In the experiments on the U250, the Sparse-Dense and Sparse-Sparse implementations outperformed the dense implementation by 56.5X and 112.3X respectively (Figure~\ref{fig:gsc1}). The increased performance delta between the dense and sparse implementations can be attributed to the relative compactness of sparsity allowing significantly more sparse networks to be accommodated on the chip (e.g. 20 Sparse-Sparse networks versus 4 dense networks). This results in the observed increase in aggregate throughput. Only one copy of each sparse network could fit on the ZU3EG, thus overall throughput on this platform was identical to that in Table~\ref{tab:single_network}.

Note that the 20X replication count achieved for the Sparse-Sparse implementation is lower than the 24X replication achieved for the Sparse-Dense. The added complexity of handling sparse activation indices (see Section~\ref{sec:sparse-routing}) increases the FPGA resources required to support the network. Nevertheless, the additional performance benefits associated with exploiting activation sparsity more than outweigh the resource costs, almost doubling the aggregate throughput.

\begin{table}[hb]
\caption{Full-Chip throughput of sparse and dense networks on the U250, measured in words processed per second. The relatively compact footprint of the sparse networks allowed the compiler to fit a larger number of networks per chip. The Sparse-Sparse implementation was over 100X faster than the Dense implementation.}
    \label{tab:full_chip}
    \begin{tabular}{ccccc}\toprule
    \textbf{FPGA Platform} & \textbf{Network Implementation} & \textbf{Total Networks} & \textbf{Throughput} & \textbf{Speedup} \\\midrule
          & Dense & 4 & 12,195 & 1.0  \\
         U250 & Sparse-Dense & 24 & 689655 & 56.5 \\
          & Sparse-Sparse & 20 & 1,369,863 & 112.3 \\\bottomrule
    \end{tabular}
    \vskip 0.1in
    
\end{table}

\subsection{Comparisons with CPU Inference Engines}
In this section we report performance gains of our sparse GSC network on a variety of widely available inference runtimes. The CPU in these experiments is a 3.0GHz 24-core Intel Xeon 8275CL processor. Figure~\ref{fig:gsc3} demonstrates the speedup of the sparse-dense network on these runtimes (relative to the dense network on the same engine) observed for our GSC CNN network. Most strikingly, both the well-known ONNX Runtime~\cite{onnx} and OpenVino~\cite{vino} runtimes fail to exploit sparsity. For the other runtime engines the sparse networks outperform the dense network, with Neural Magic's Deepsparse \cite{magic} and the Apache TVM providing a 2X and 3X speedup, respectively. The observed performance gains are relatively modest, considering there is a 20X reduction in the number of non-zero weights.

In Figure~\ref{fig:gsc4}, the absolute performance for the sparse networks on the CPU and FPGA are compared. The results show significant speedups from sparsity on an FPGA with absolute performance over 10X that currently achievable on a CPU system. None of these runtime engines exploit both sparsity in activations and weights.

\subsection{Power Efficiency}
\begin{table}[]
\begin{center}
\caption{Power efficiency of sparse networks on the U250 and ZE3EG FPGAs in comparison with the dense network baseline. We estimate power efficiency using a word/sec/watt metric based on worst-case (i.e. total system power of each platform).}
\label{tab:power}
\begin{tabular}{cccccc}
\toprule
\textbf{\shortstack{FPGA\\Platform}} & \textbf{\shortstack{System\\Power (W)}} & \textbf{\shortstack{Network\\Type}} & \textbf{\shortstack{Number of\\Networks}} & \textbf{\shortstack{Words\\Sec/Watt}} & \textbf{\shortstack{Relative\\Efficiency, \%}} \\
\midrule
 &  & Dense & 4 & 54 & 100 \\
 &  & Sparse-Dense & 1 & 158 & 292 \\
U250 & 225 & Sparse-Dense & 24 & 3065 & 5675\\ 
 &  & Sparse-Sparse & 1 & 455 & 842 \\
 &  & Sparse-Sparse & 20 & 6088 & 11274\\ 
\\
 &  & Dense & 0 & 0 & 0\\
ZU3EG & 24 & Sparse-Dense & 1 & 877 & 1624\\
 &  & Sparse-Sparse & 1 & 1893 & 3505\\\bottomrule
\end{tabular}
\end{center}
\end{table}

In addition to improved inference performance, reduced power consumption is becoming increasingly critical \citep{carbon, Thomson2020}. Table \ref{tab:power} shows the absolute and relative power efficiency for inference operations. Due to the significant resource reductions associated with sparse networks, not only has the total throughput improved, but the power consumed per inference operation has also dropped considerably. It is common to improve throughput at the expense of increasing energy consumption \citep{Thomson2020}. Our results demonstrate that sparsity avoids many unnecessary operations altogether, simultaneously improving throughput and power efficiency.

\section{Resource Tradeoffs Analysis}
\label{sec:resnet_results}
In the previous section, we discussed end-to-end throughput results for a full network. It became clear during implementation that a key consideration is the resource usage required to implement the routing and $k$-WTA components. In this section we implemented a series of controlled experiments to analyze these resource tradeoffs in isolation. 

\begin{figure}[t]
    \centering
    \includegraphics[width=0.8\textwidth]{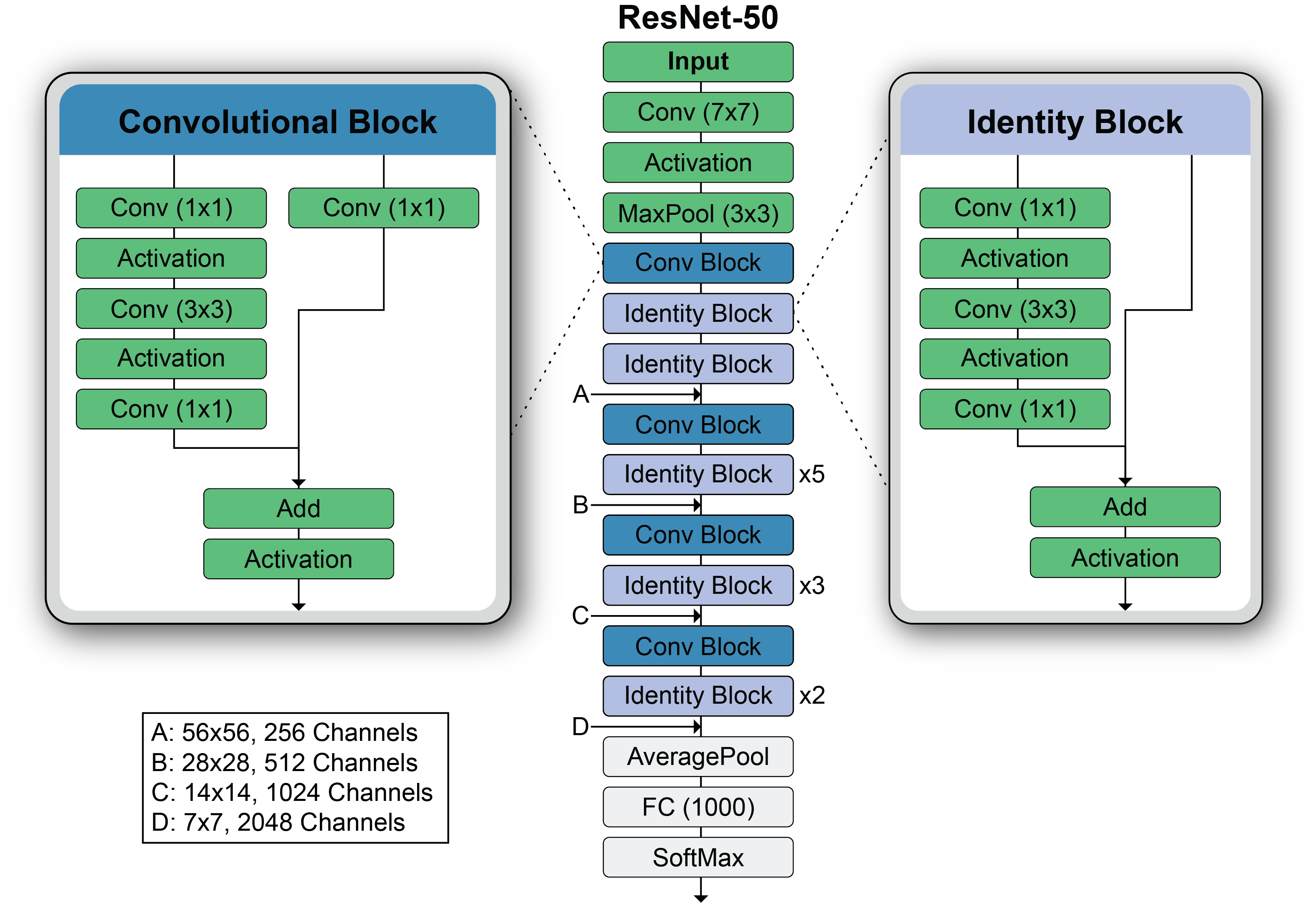}
    \caption{Overview of a ResNet-50 architecture, illustrating the repeated use of identity and convolutional blocks, and the increasing number of channels deeper in the network. As can be seen, most of the layers use either $1 \times 1$ or $3 \times 3$ kernel sizes. The very first "stem" layer uses a $7 \times 7$ kernel size.}
    \label{fig:resnet}
\end{figure}

In GSC, the convolutional layers employed $5\times5$ kernels. In these experiments we focus on two other structures, $1 \times 1$ and $3 \times 3$ kernel types. These kernel types are typical of a number of common networks structures, such as the ResNet-50 (Figure~\ref{fig:resnet}), ResNeXt, and MobileNetV2 networks \citep{He2015a, Xie2017, Sandler2018}. We investigate the resource savings achievable via a combination of activation and weight sparsity applied to these convolutional layer types. The key questions revolve around how the FPGA resource requirements scale with weight sparsity, and how this changes as we add in activation sparsity.

\subsection{Experiment Setup}
\label{sec:res_arch}
To investigate whether Complementary Sparsity could be applied generally, we developed the component shown in Figure~\ref{fig:spsp} as a set of general-purpose parameterized blocks. For the $k$-WTA block, $K$ is defined per instance at compilation time. Three convolutional blocks were developed: a sparse-dense 7$\times$7 convolutional block, and separate blocks for $1\times1$ and $3\times3$ sparse-sparse convolutions. The parameterization of these blocks included: boundary padding size, stride, weight sparsity, and memory bandwidth, as well as input activation sparsity for the 1$\times$1 and $3\times3$ blocks.

When implementing components on an FPGA there is a great deal of flexibility in choosing how resources are allocated. There is significant latitude to trade serial processing for parallel processing by allocating sufficient resources to every stage. This in turn makes it challenging to explore both resource utilization and throughput in a controlled manner. In these implementations, we targeted a fixed throughput for all components in order to focus on resources. Our throughput target was chosen to be aggressive without leading to exploding resources. The primary target stipulated that a 1$\times$1 [64:64]\footnote{Our notation [$a$:$b$] refers to an input channel count of $a$ and an output channel count of $b$.} convolution should be computed in a single cycle. For a 1x1 [64:64] convolution, when weights and activations are dense, 4096 multiplications and 4096 additions are required to carry out the computation per spatial location. For a sparse-sparse computation ($N$=4 and $K$=8), this requirement is reduced to 32 multiplications and 32 additions, making this aggressive target feasible. Our $3\times3$ [64:64] convolution used nine 1$\times$1 convolutions, taking about 9 cycles. The $k$-WTA layer had a target of one cycle. As sparsity levels varied the compiler automatically allocated the hardware resources to achieve this target, allowing a controlled investigation of resource impact. We removed bandwidth as a confounding parameter by allocating sufficient memory to meet the target (but see Section~\ref{sec:bandwidth} below for an analysis of bandwidth). 

The parameterization and above setup facilitated a systematic analysis of Complementary Sparsity for a variety of convolutional layers, primarily as a function of weight and activation sparsities. Our goal was to gain an improved understanding of the resource consumption and its scaling with degree of sparsity. Since extending sparse architectures to dense configurations is not meaningful, we confine our analysis to $k$-WTA activation sparsity $\geq 50\%$, and weight sparsity $\geq 50\%$. These represent reasonable break points between sparse and dense implementations.

\begin{figure}[!tbp]
  \subfloat[]{\includegraphics[width=0.5\textwidth]{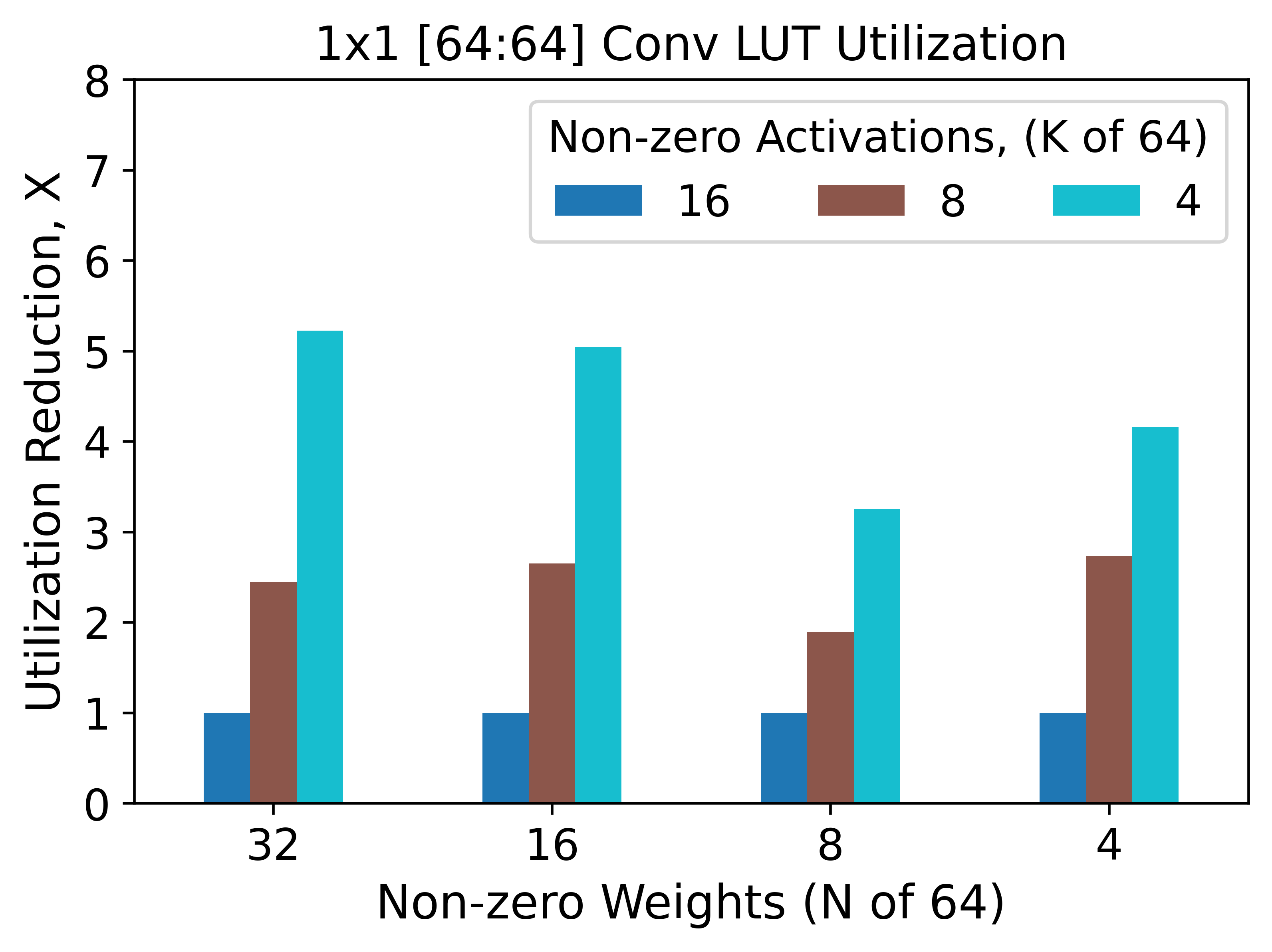}\label{fig:1x1_LUT}}
  \subfloat[]{\includegraphics[width=0.5\textwidth]{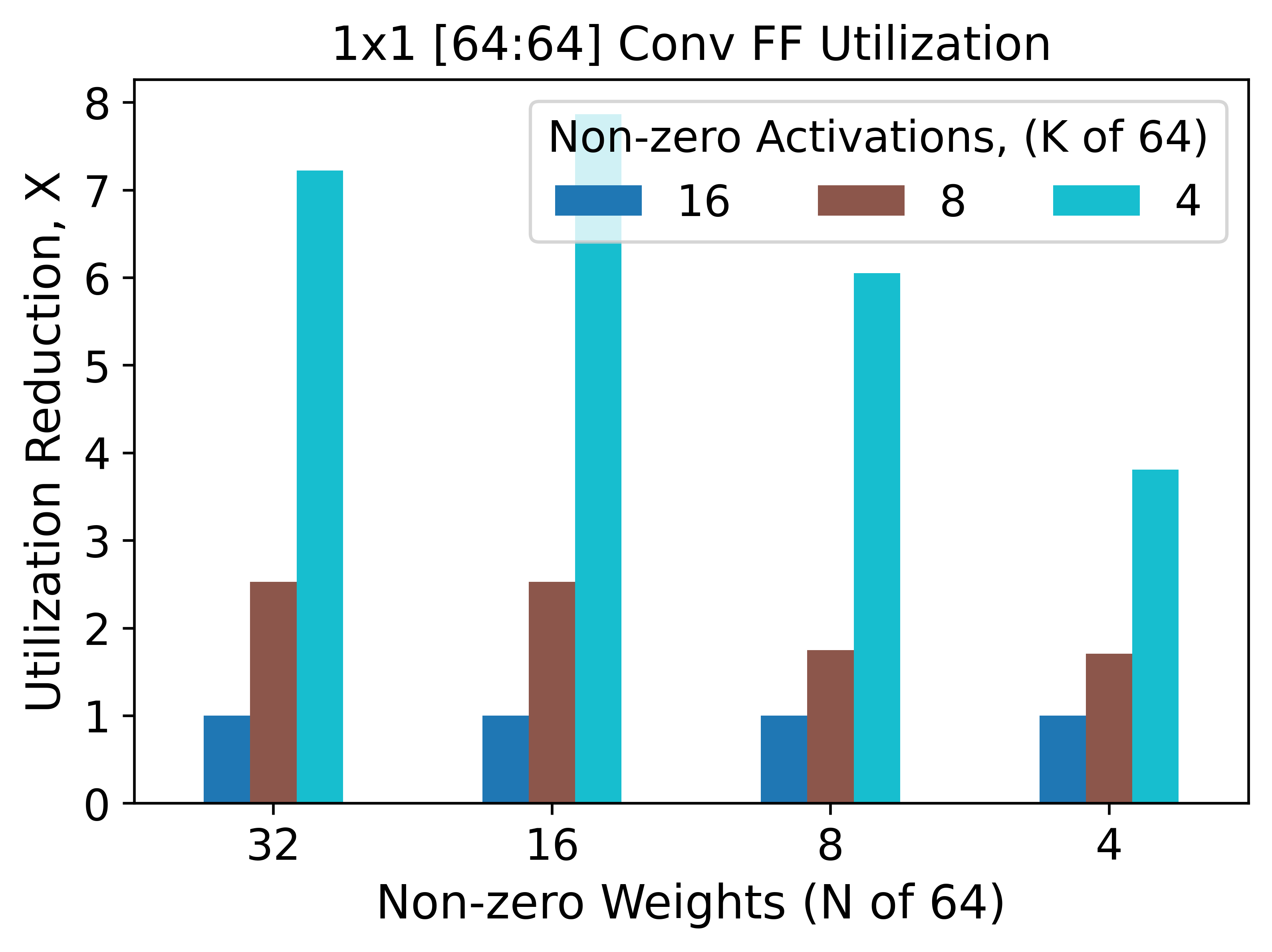}\label{fig:1x1_FF}}
  \\
  \centering
    \subfloat[]{\includegraphics[width=0.5\textwidth]{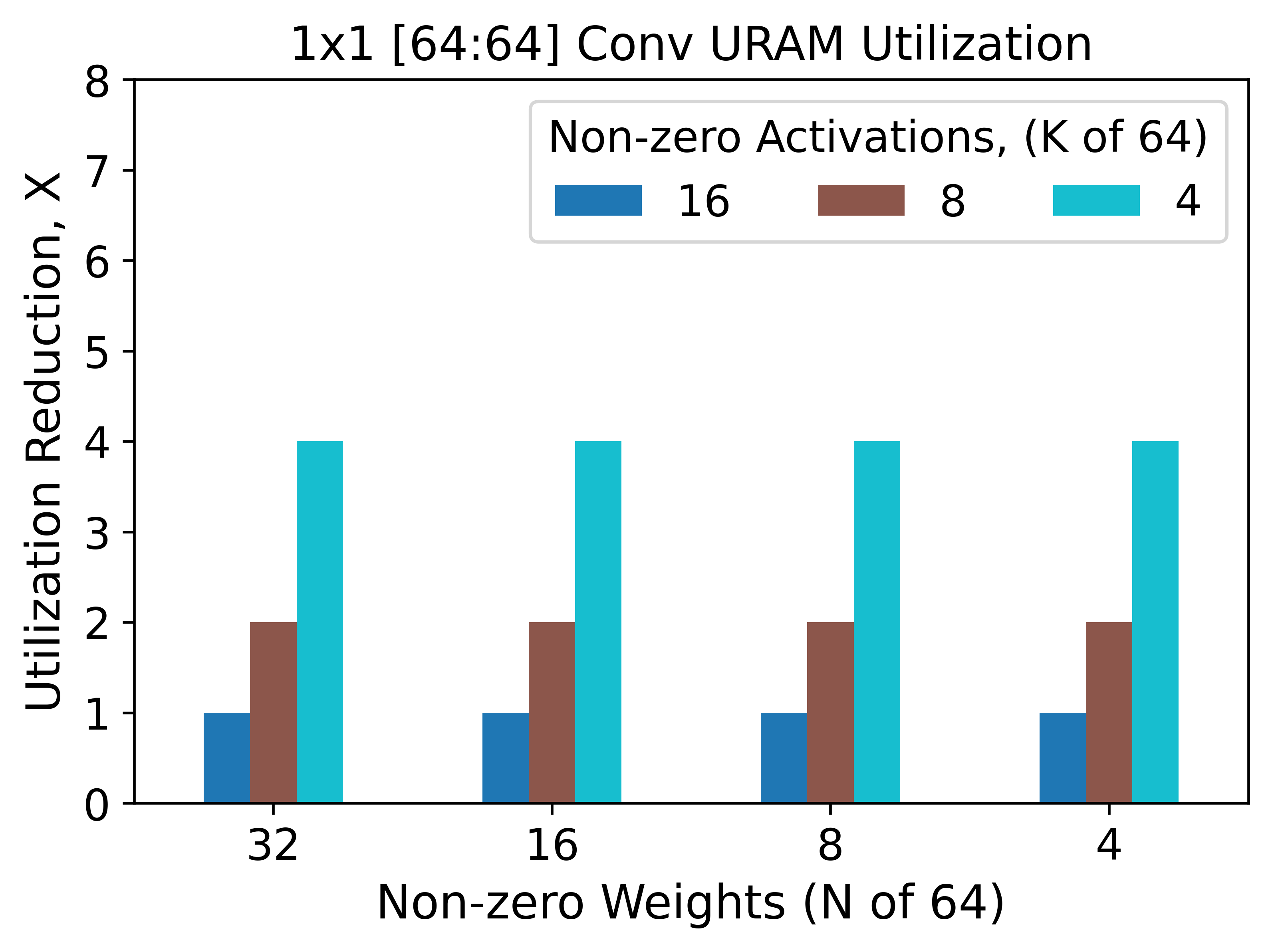}\label{fig:1x1_RAM}}
  \caption{Impact of activation sparsity on resource utilization for 1$\times$1 [64:64] convolution operations for different degrees of weight sparsity for: (a) LUTs, (b) FFs and (c) URAMs (K and N indicate the number of non-zero elements, reduction in utilization relative to K=16).}
\end{figure}

\subsection{Resource Utilization of Sparse-Sparse Convolution Kernels}
\begin{figure}[!tbp]
  \subfloat[]{\includegraphics[width=0.5\textwidth]{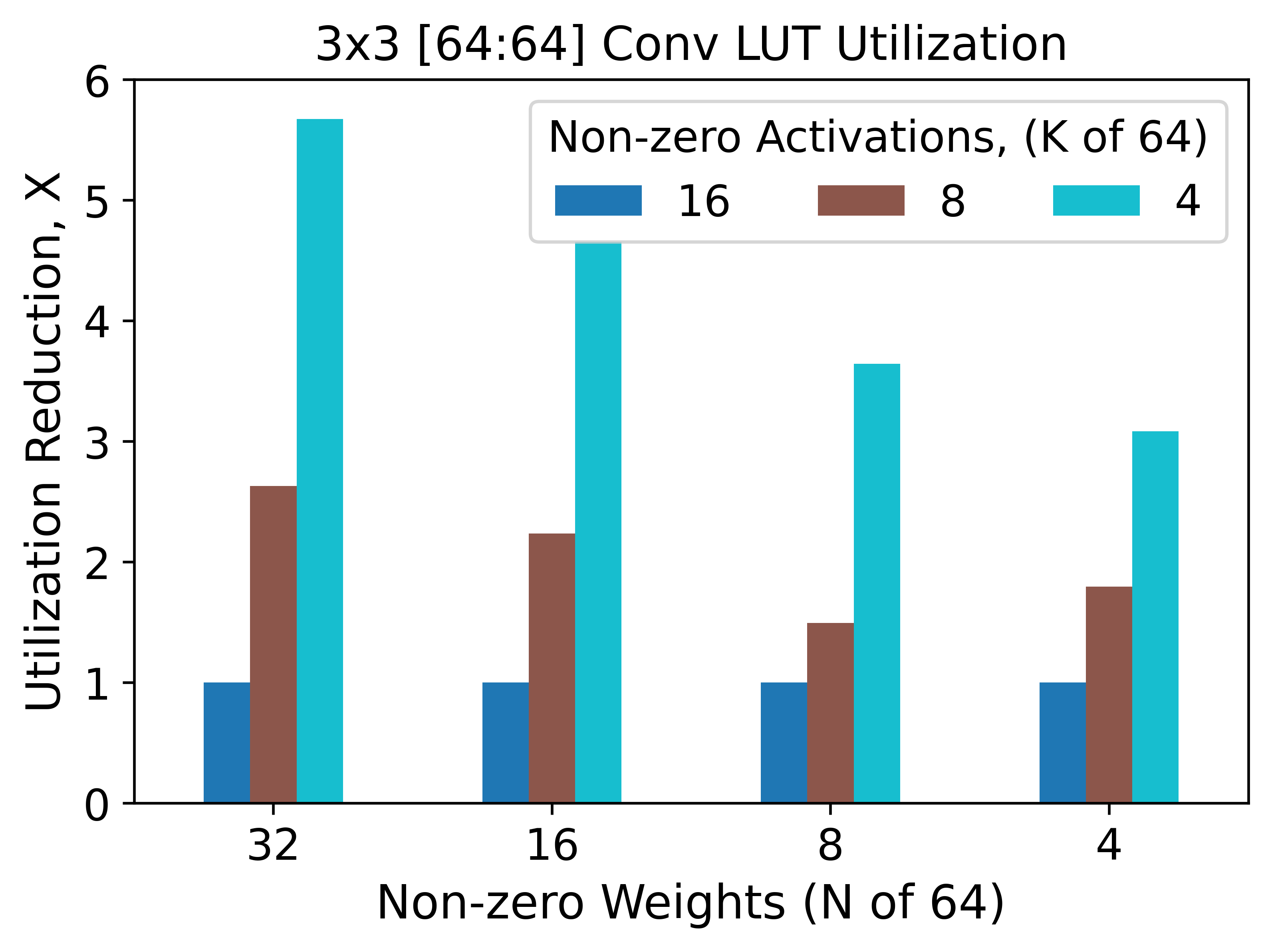}\label{fig:3x3_LUT}}
  \subfloat[]{\includegraphics[width=0.5\textwidth]{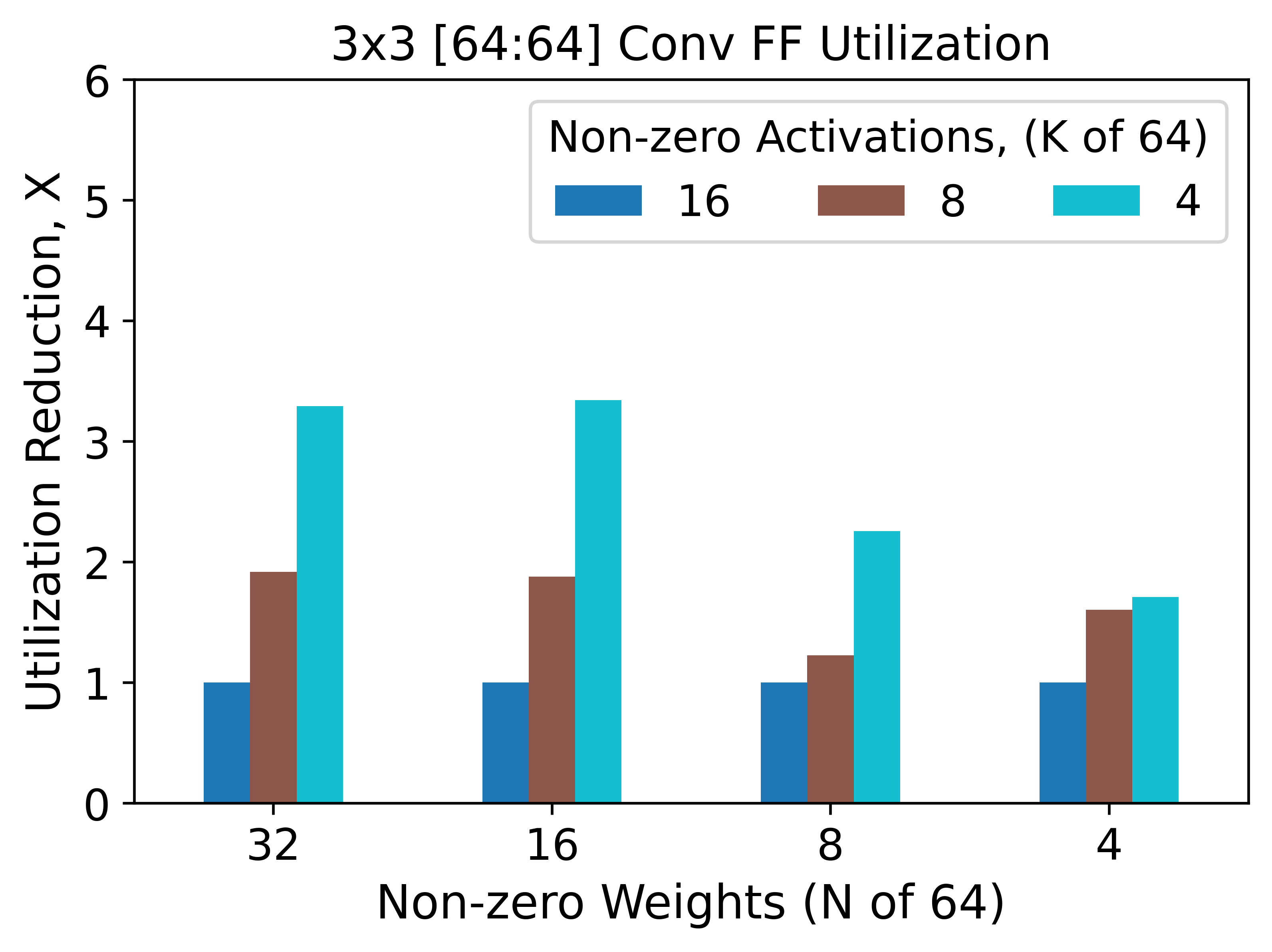}\label{fig:3x3_FF}}
  \\
  \centering
    \subfloat[]{\includegraphics[width=0.5\textwidth]{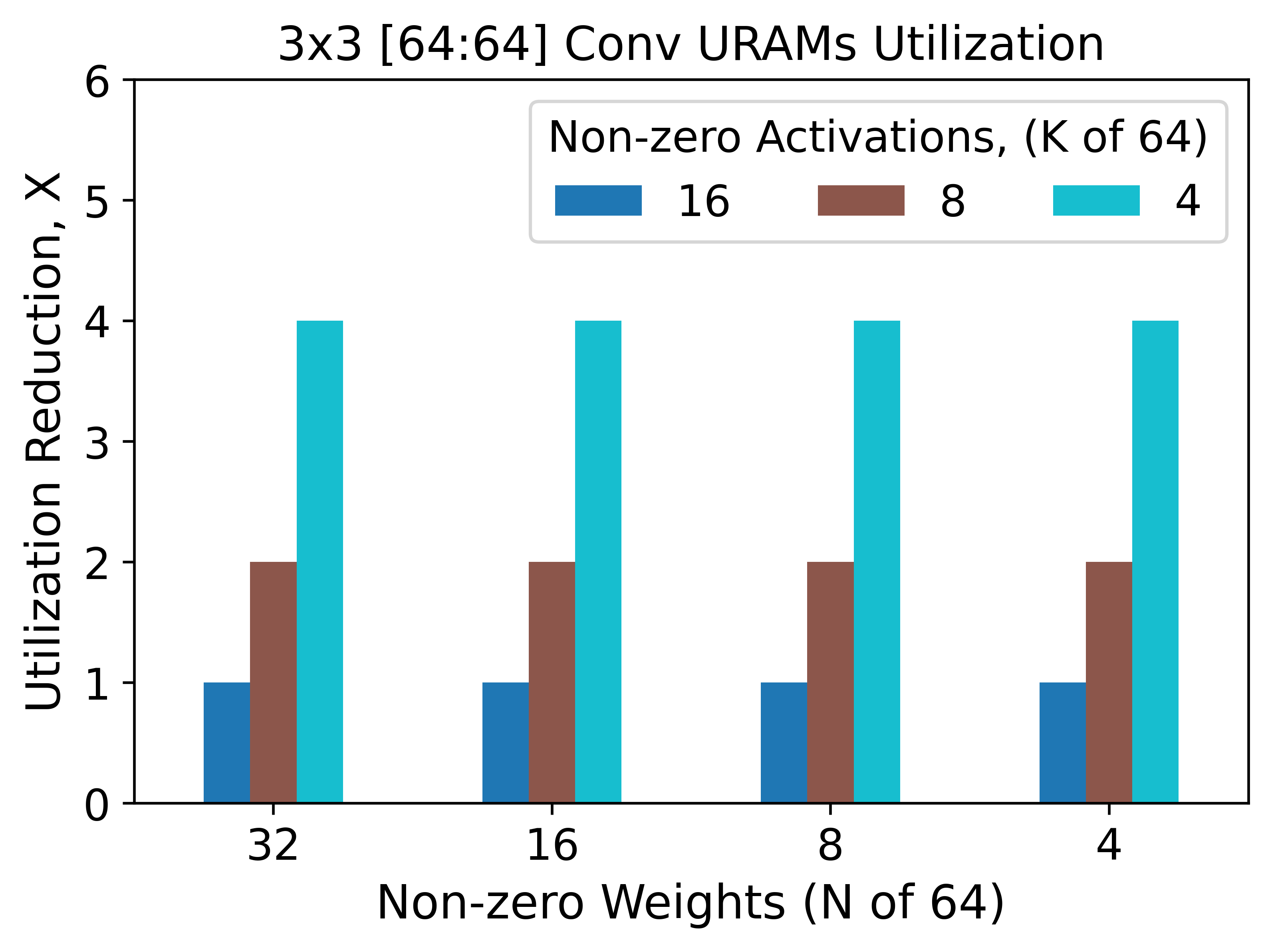}\label{fig:3x3_RAM}}
  \caption{Impact of activation sparsity on resource utilization for 3$\times$3 [64:64] convolution operations for different degrees of weight sparsity for: (a) LUTs, (b) FFs, (c) URAMs (K and N indicate the number of non-zero elements, reduction in utilization relative to K=16).}
\end{figure}

In Figure~\ref{fig:1x1_LUT}-\ref{fig:1x1_RAM}, and Figure~\ref{fig:3x3_LUT}-\ref{fig:3x3_RAM}, we present the resource utilization observed for the convolutional layers when activation sparsity is increased. In each experiment, we hold the weight sparsity constant, increase the activation sparsity and report the reduction in resource utilization. Our FPGA implementations of convolutional layers consume a variety of FPGA resources, including Lookup Tables (LUTs), Flip Flops (FFs), and memory blocks (URAMs). We found that, for all investigated levels of weight sparsity, increasing the activation sparsity delivered a significant reduction in the resource utilization across all FPGA resources. For example, looking a Figure~\ref{fig:1x1_LUT}, for a weight sparsity of 4 non-zeros out of 64-elements (i.e. $\frac{60}{64}$=93.75\% sparse), as the activation sparsity is increased from $\frac{16}{64}$ to $\frac{8}{64}$ and $\frac{4}{64}$, the number of LUTs required for the implementation of the 1$\times$1 convolution is reduced by 2.7X and 4.1X, respectively.

It is apparent that, across all levels of weight sparsity investigated, as the level of activation sparsity is increased, the complexity of the implementations of the convolutional layers is reduced. The degree to which the resource utilization decreased with increased sparsity is dictated by the resource type and the degree of weight sparsity. For instance, there is a clear linear relationship between URAMs consumed and the degree of activation sparsity. This is true for all levels of weight sparsity investigated.

LUTs are used for both routing and implementing multipliers in this design. The LUT count is reduced as a function of weight sparsity due to the decreased number of multipliers, while conversely there is greater routing complexity with increased weight sparsity. This latter effect is due to managing larger numbers of consolidated sparse weight kernels. Despite these competing factors, the overall LUT count decreases significantly with weight sparsity. 

For FF utilization, the resource savings are more muted at higher weight sparsities. For FFs, which are primarily used for high bandwidth local storage, there is a baseline quantity for holding input and output values. The especially muted 3$\times3$ convolution FF resource utilization results illustrated in Figure~\ref{fig:3x3_FF} reflect the fact that FFs are also used to buffer intermediate results; the results of the 9 internal $1\times1$ operations are serially accumulated. Therefore the FF utilization scaling as a function of weight sparsity is on top of these relatively static baselines. However, in many instances we demonstrate a super-linear reduction in resource utilization. This is rate of reduction is observed because a number of the elements in the implementations of the convolutions scale non-linearly with the number of non-zero activations; resulting in significant resource savings as $K$ is decreased.

\begin{figure}[!tbp]
  \subfloat[]{\includegraphics[width=0.5\textwidth]{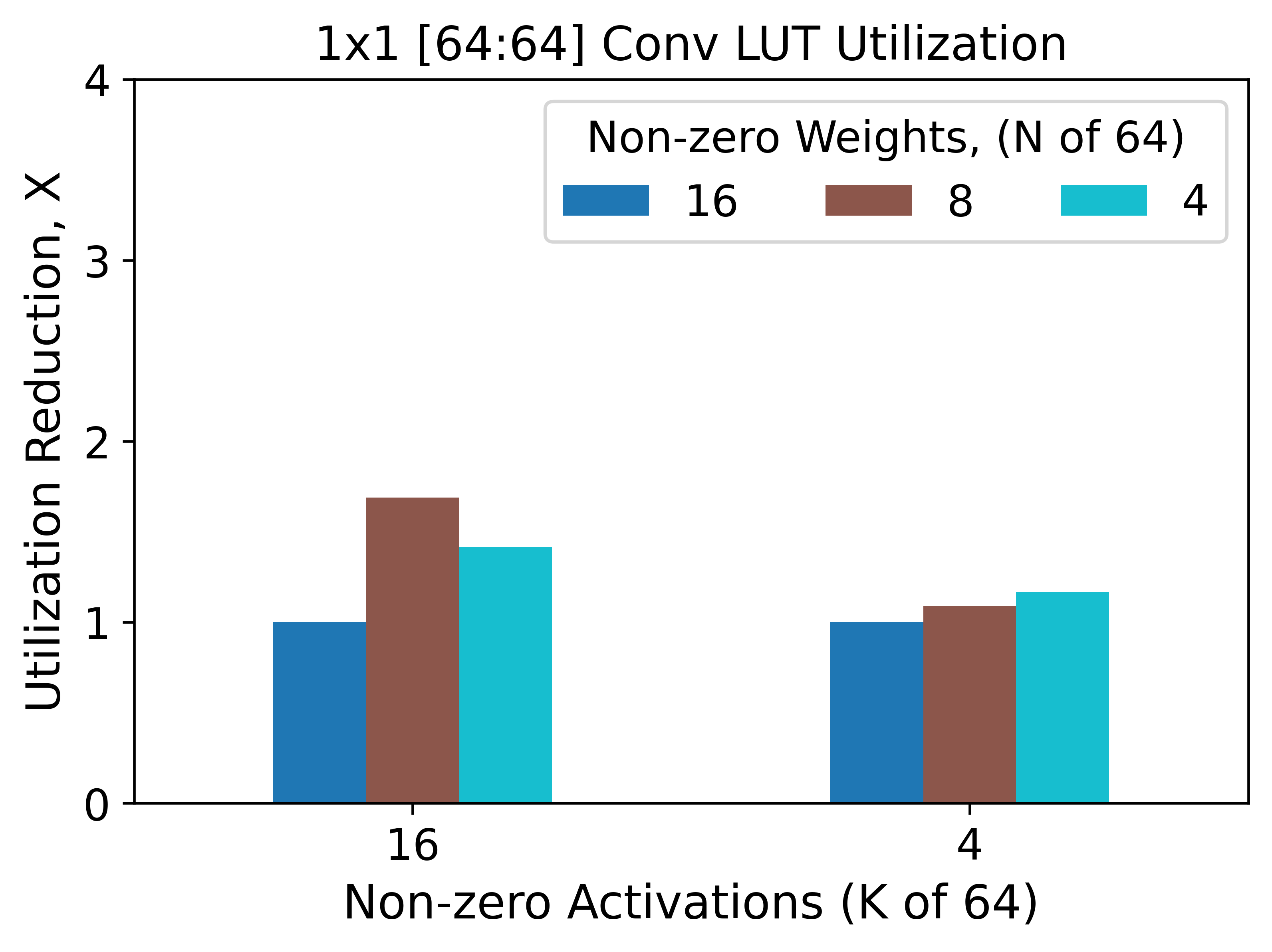}\label{fig:1x1_LUTw}}
  \subfloat[]{\includegraphics[width=0.5\textwidth]{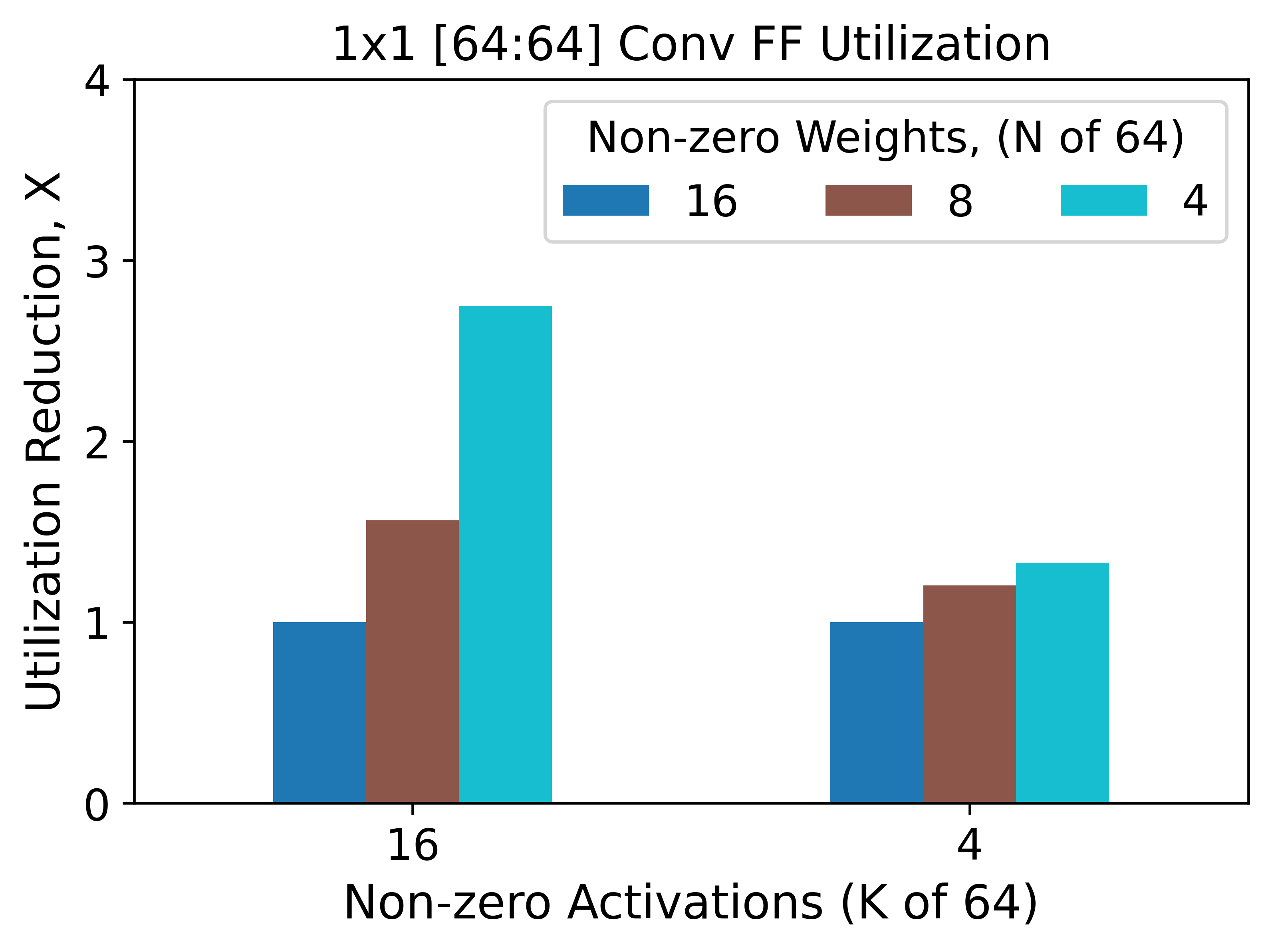}\label{fig:1x1_FFw}}
  \\
  \centering
    \subfloat[]{\includegraphics[width=0.5\textwidth]{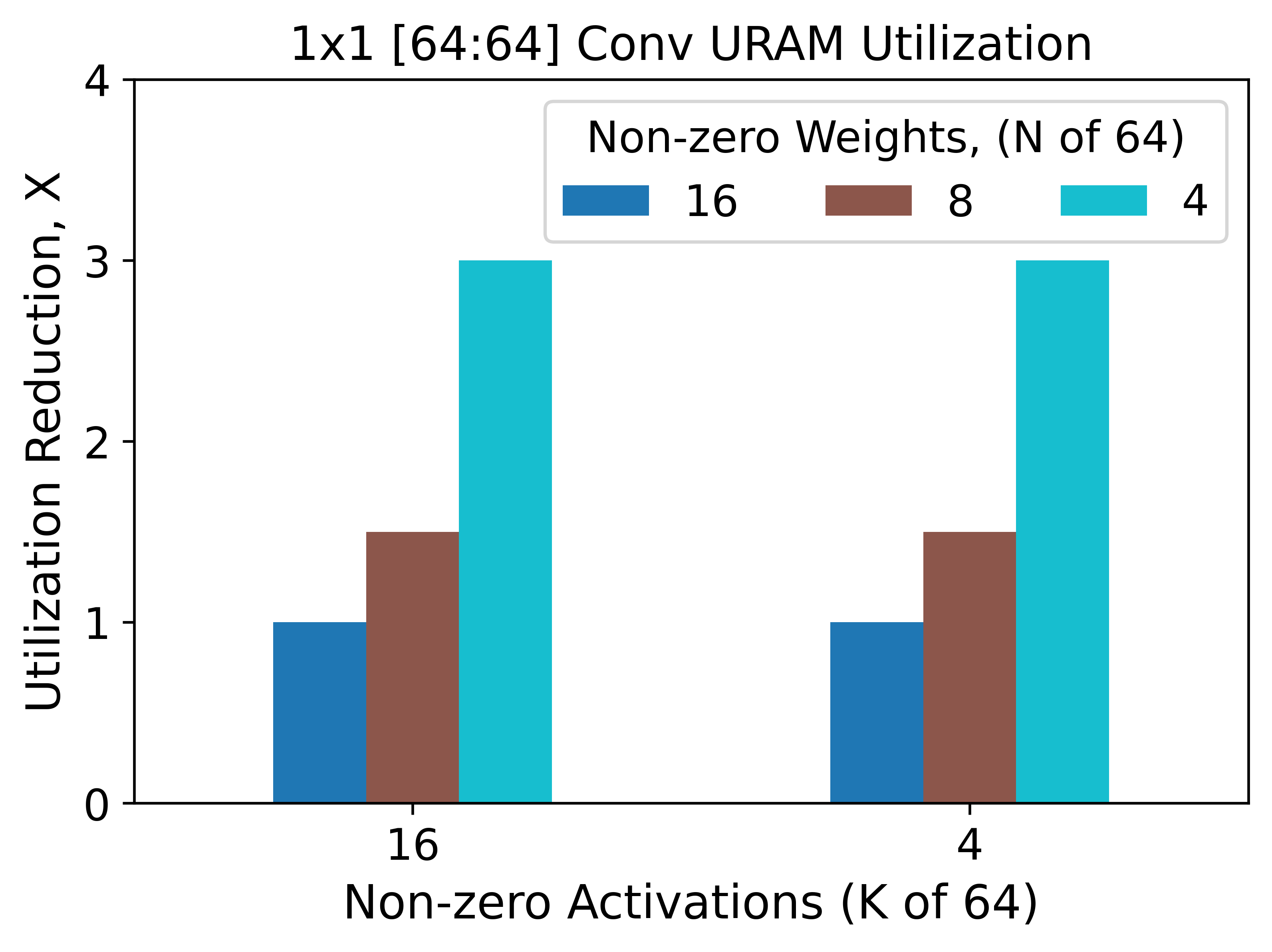}\label{fig:1x1_RAMw}}
  \caption{Impact of weight sparsity on resource utilization for 1$\times$1 [64:64] convolution for different degrees of activation sparsity for: (a) LUTs, (b) FFs and (c) URAMs (K and N indicates the number of non-zero elements, reduction in utilization relative to K=16).}
\end{figure}

\begin{figure}[!tbp]
  \subfloat[]{\includegraphics[width=0.45\textwidth]{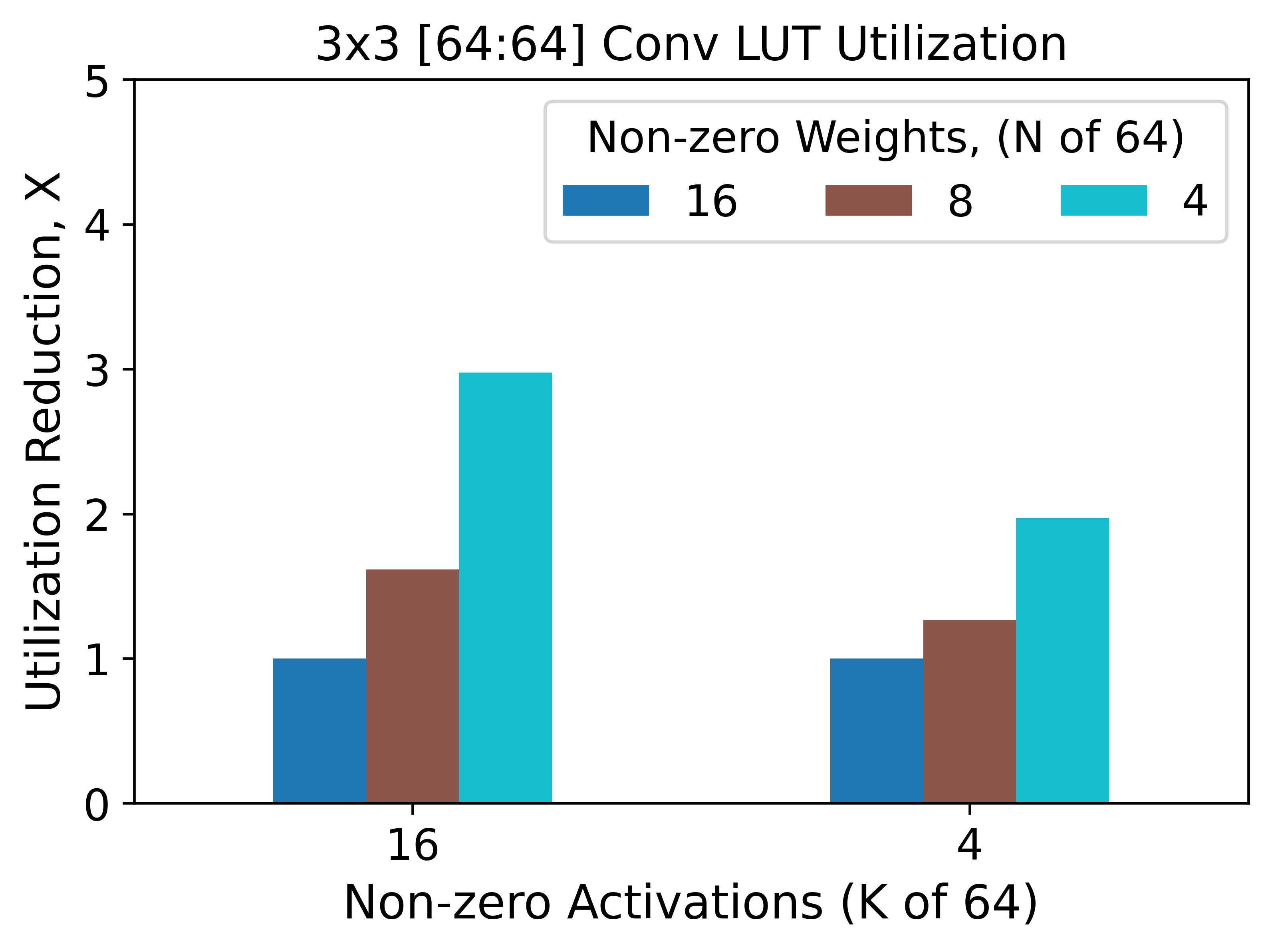}\label{fig:3x3_LUTw}}
  \subfloat[]{\includegraphics[width=0.45\textwidth]{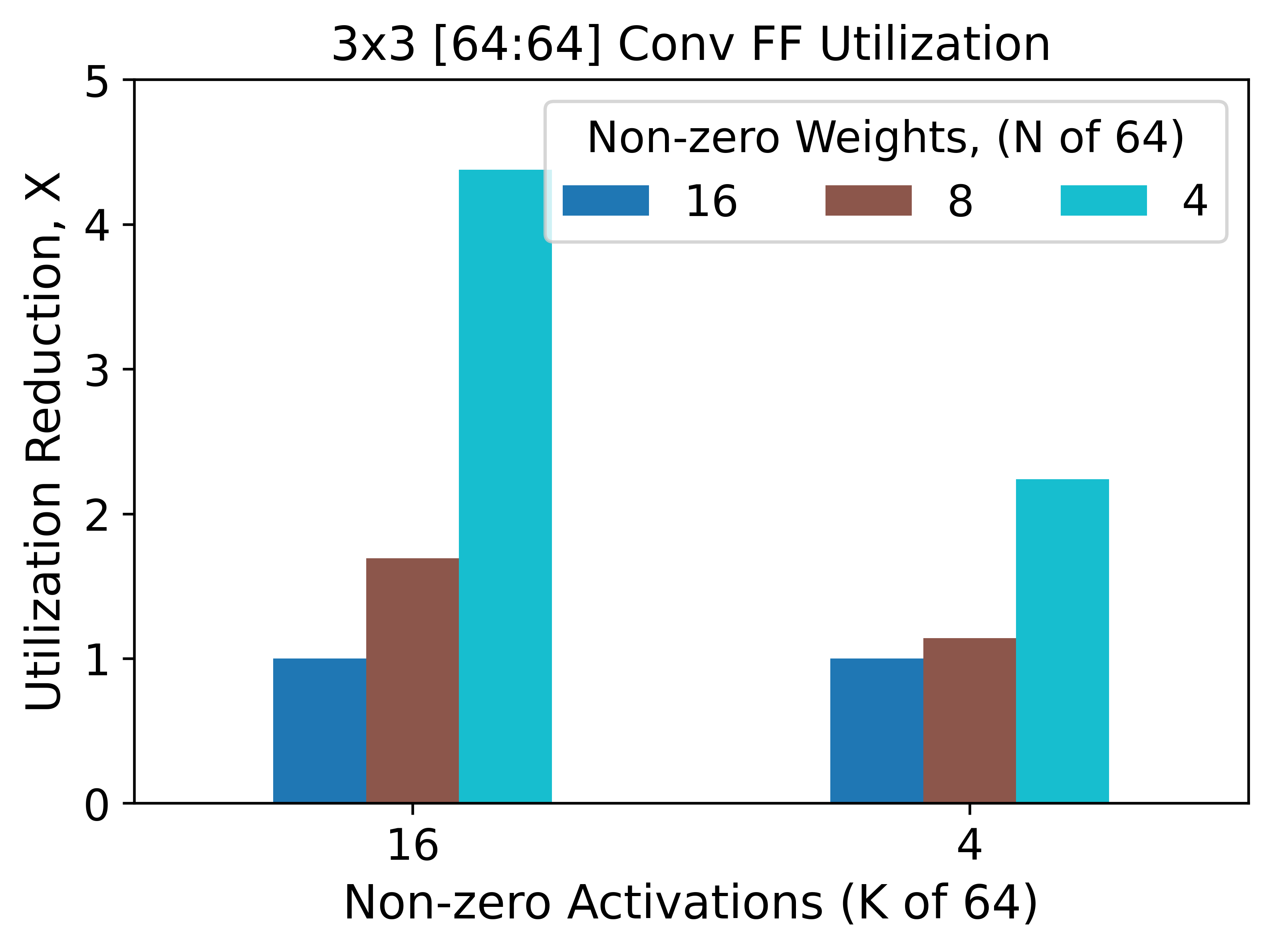}\label{fig:3x3_FFw}}
  \\
  \centering
    \subfloat[]{\includegraphics[width=0.45\textwidth]{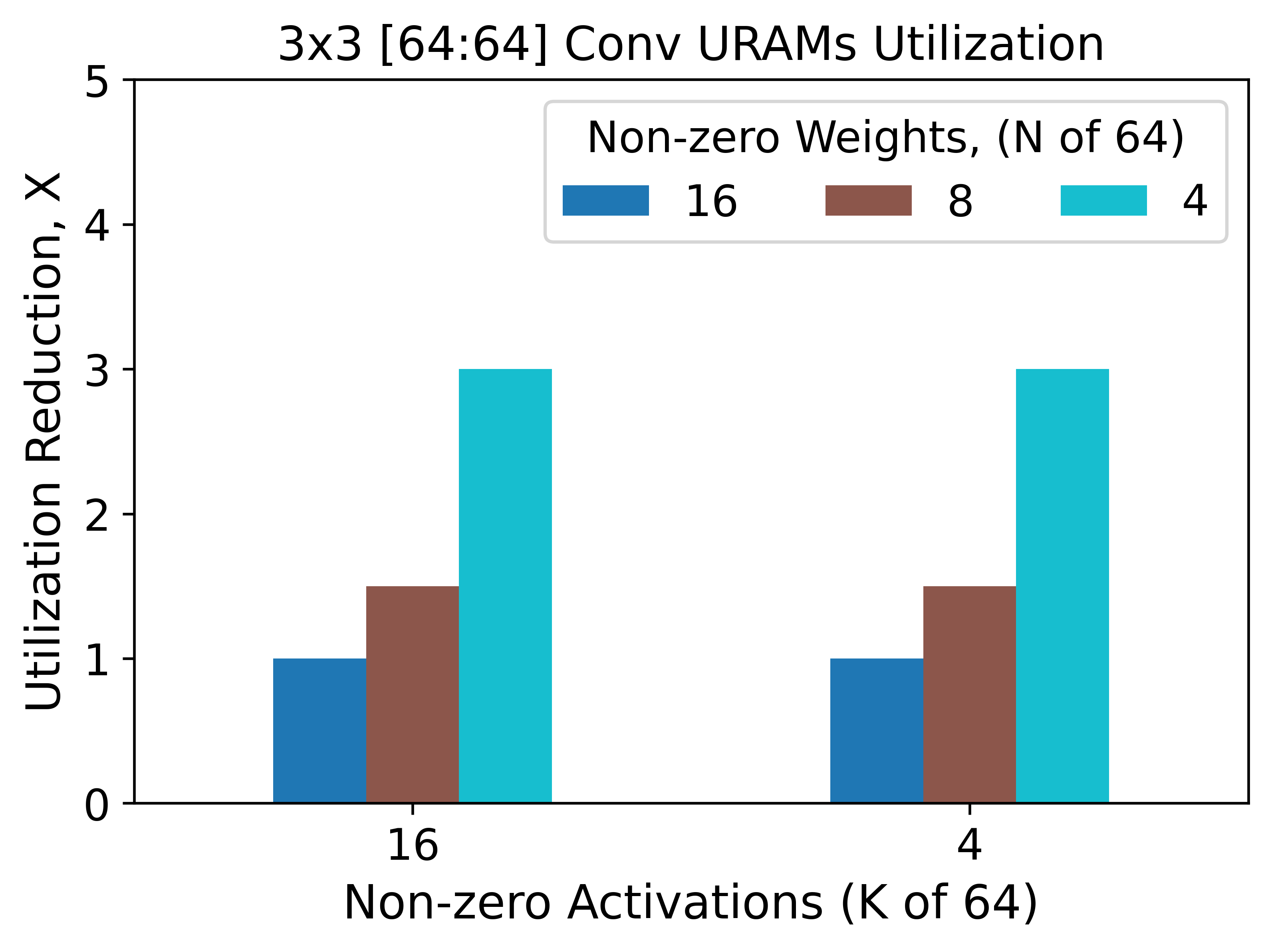}\label{fig:3x3_RAMw}}
  \caption{Impact of weight sparsity on resource utilization for 3$\times$3 [64:64] convolution operations for different degrees of activation sparsity for: (a) LUTs, (b) FFs, (c) URAMs (K and N indicate the number of non-zero elements, reduction in utilization relative to N=16).}
\end{figure}

Figure~\ref{fig:1x1_LUTw}-\ref{fig:1x1_RAMw} and Figure~\ref{fig:3x3_LUTw}-\ref{fig:3x3_RAMw} show the impact of varying weight sparsity for a fixed activation sparsity. In these results, the resource savings are sub-linear. Increases in weight sparsity result in decreases in the number of multiplies, but as discussed, routing overheads limit these reductions. However, for LUTs, FFs and URAMs, at any given activation sparsity, increasing the weight sparsity reduces the resource consumed by the implementation. 

In summary, for our FPGA implementations using Complementary Sparsity, increasing sparsity (weight, activation or both) results in more resource efficient implementations, while continuing to meet the stipulated throughput metric.

\subsection{Resource Utilization of $k$-WTA}
\label{sec:kwta-resou}
\begin{figure}
    \centering
    \includegraphics[width=0.5\textwidth]{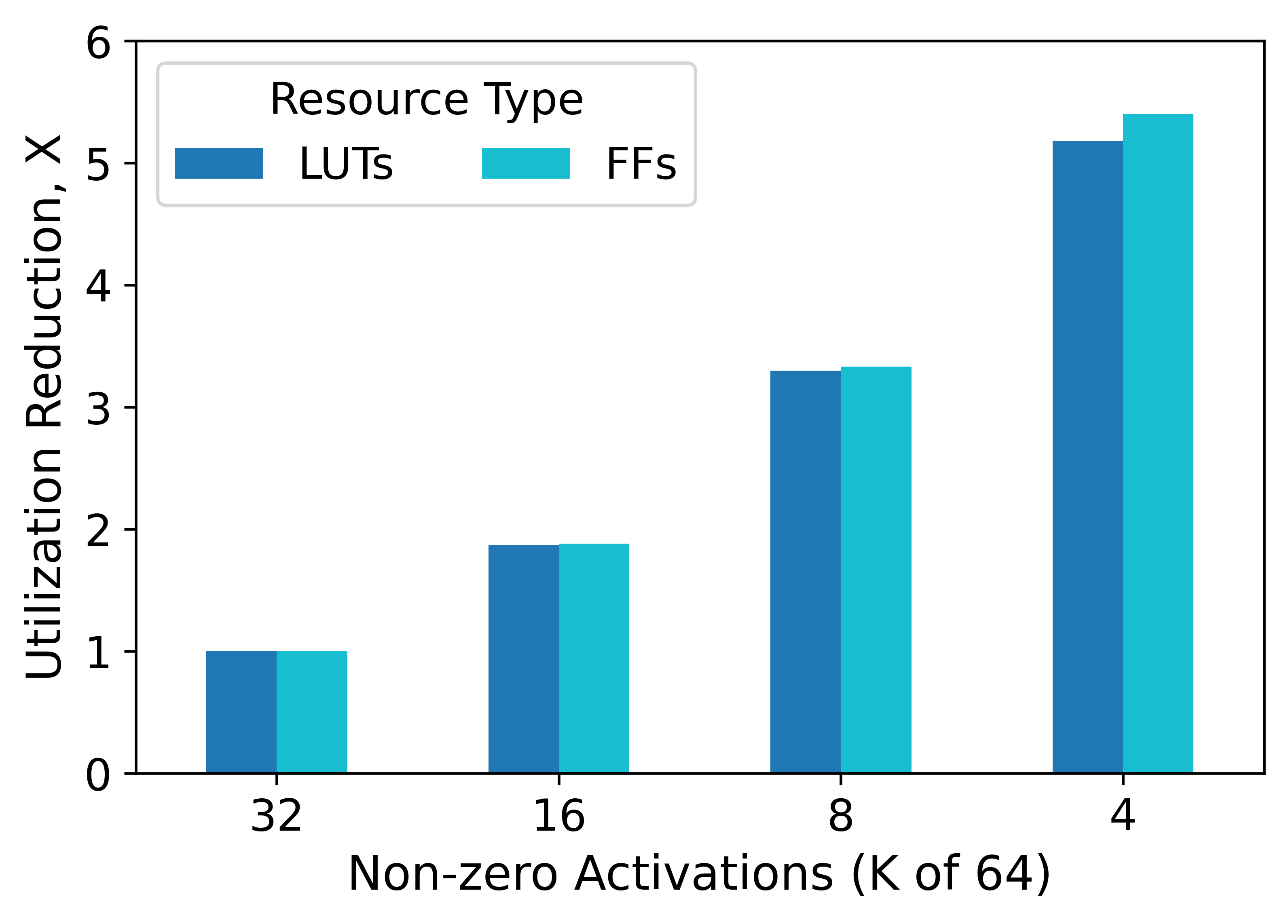}
    \caption{Impact of activation sparsity levels on $k$-WTA resource utilization (K indicates the number of non-zero elements, reduction in utilization relative to K=32).}
    \label{fig:kwta}
\end{figure}

We also investigated the resource impact of our $k$-WTA implementations. Here too increasing activation sparsity resulted in the consumption of fewer hardware resources, as illustrated in Figure~\ref{fig:kwta}. The resource utilization was found to decrease almost linearly with the degree of sparsity. This represents an important synergy, with the convolutional kernel implementations benefiting from increased levels of activation sparsity, and the cost of providing the sparse activations decreasing as the level of activation sparsity is increased.

\begin{figure}[!tbp]
  \subfloat[1x1 Convolution + $k$-WTA]{\includegraphics[width=0.5\textwidth]{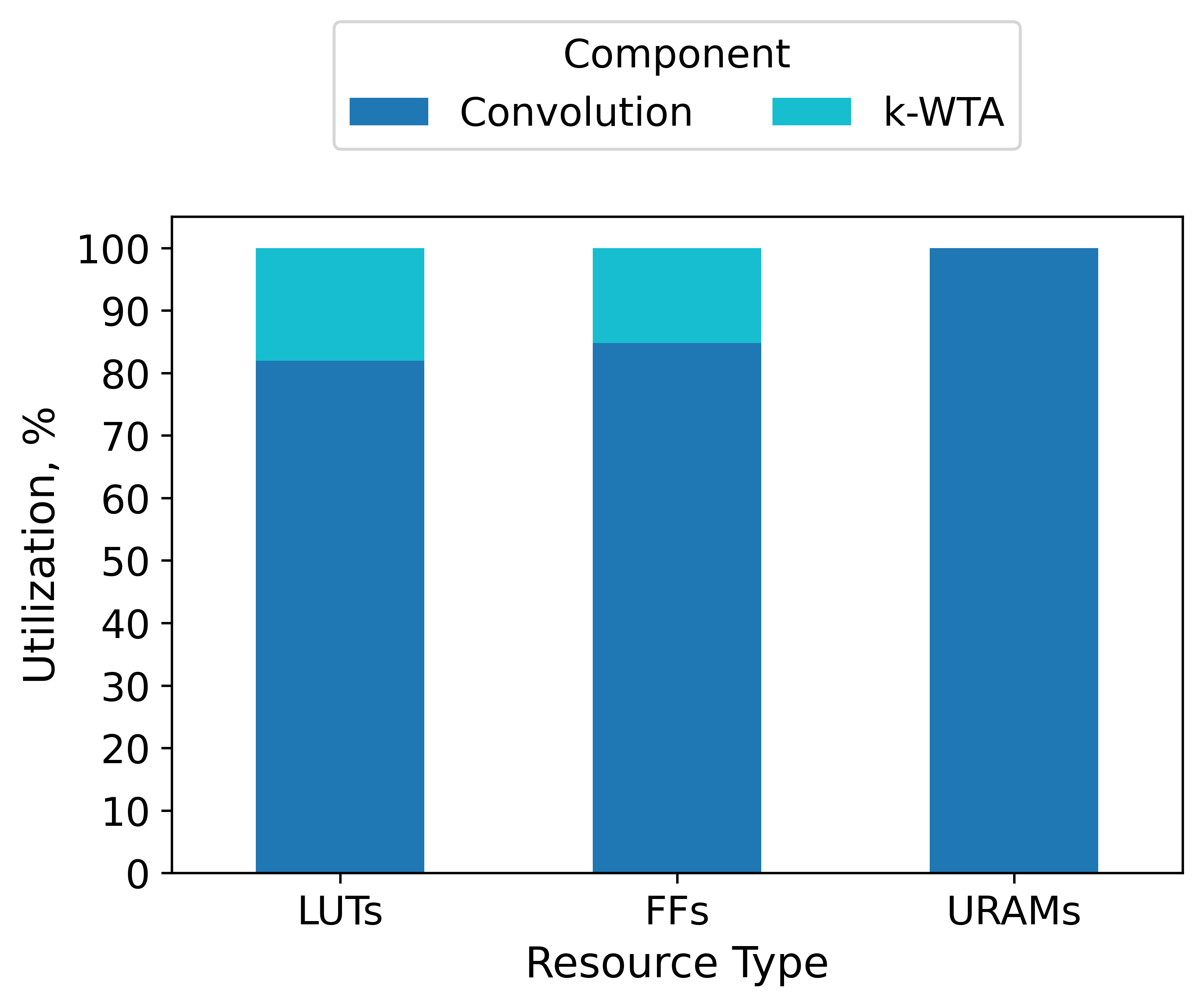}\label{fig:1x1_total}}
  \subfloat[3x3 Convolution + $k$-WTA]{\includegraphics[width=0.5\textwidth]{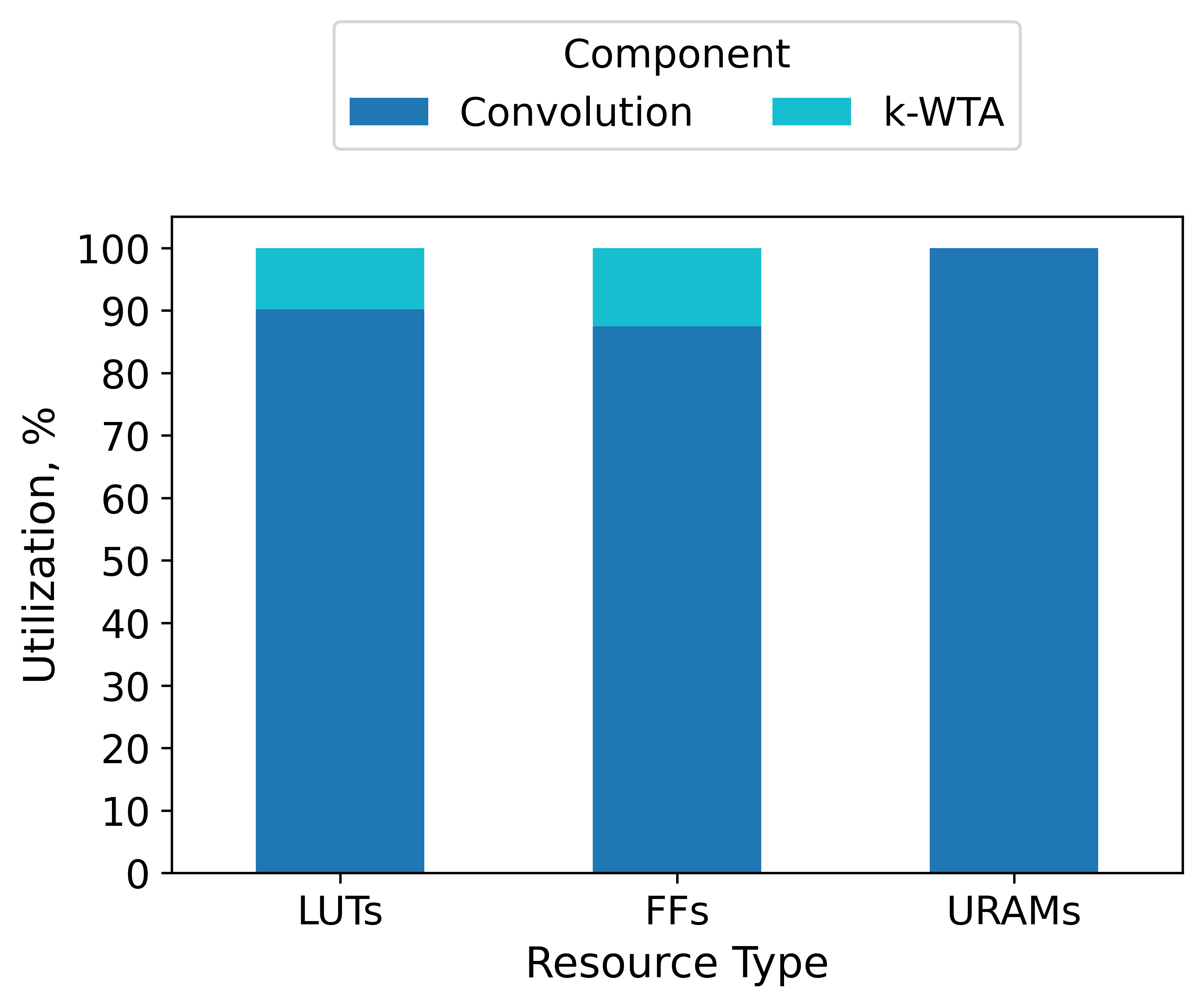}\label{fig:3x3_total}}
  \caption{Total resource utilization of convolution operations and associated $k$-WTA components (for N=8 and k=8). $k$-WTA consumes a relatively small percentage of LUTs and FFs, and no URAMS.}
\end{figure}

In Figures~\ref{fig:3x3_total}-~\ref{fig:1x1_total}, the combined resource utilization for sparse-sparse convolutions and their associated $k$-WTA components is shown. For both the $1\times1$ and $3\times3$ convolutions, the costs associated with the $k$-WTA implement is small compared with the costs associated with the convolutions, especially for the $3\times3$ convolution, where the implementation cost of the convolution is increased, but the $k$-WTA cost remains constant.

\subsection{Sparsity in the Network Stem}
\label{sec:stem}
In addition to the convolutional kernels that form the convolutional and identity blocks in ResNet-50, the network contains an initial ``stem''. This stem performs a 7$\times$7$\times$3 (RGB color values) convolution on the input image \citep{pmlr-v119-evci20a}. In many sparse implementations, this first convolutional layer is left as a standard dense operation, because it represents a small part of the overall implementation profile. However, the implementation of this initial convolution can both require significant hardware resources and dictate overall network throughput. Although the overall latency of the entire network pipeline will shrink when recast as a sparse implementation, the throughput benefits will be capped by this first layer, making an efficient sparse implementation highly desirable.

Complementary sparsity can be successfully applied to this stem convolution, but, because the inputs to this first layer are dense images, a sparse-sparse implementation is not feasible. However, a weight-sparse implementation on an FPGA provides a considerable performance benefit: in our implementations, by increasing the weight sparsity (from N=9 to N=5) by 1.8X, we increased throughput by 1.6X. In this layer we chose to implement Complementary Sparsity in the spatial dimensions. We also imposed block sparsity constraints, with the 3-element input dimension being treated as a block, either fully non-zero or completely zero.

The first layer of most DNNs process a dense input data stream and will only be able to exploit weight sparsity in a sparse-dense configuration. If the rest of the DNN is implemented as sparse-sparse layers those layers will see large performance gains. As an unexpected result, in pipelined implementations we find that the first layer's throughput will often dictate the maximum throughput of the network. To increase overall throughput, in FPGAs it is possible to increase the parallelism of the first layer, such that its sparse-dense layer latency is less than or equal to the highest latency sparse-sparse layer. This additional resource cost is made up by the resource gains achieved in the rest of the network. As a general rule, we find that the large gains achieved by a sparse-sparse implementation warrant careful profiling of the rest of the system as unexpected bottlenecks can emerge.

\subsection{Sparse-Sparse Memory Bandwidth Considerations}
\label{sec:bandwidth}
Memory represents a scarce resource, and its efficient utilization is a key contributor to the success of sparse-sparse implementations. Two factors dictate memory utilization on the FPGA. The first is simply dictated by the capacity required to retain the relevant weight elements. Second is the requirement for sufficient memory bandwidth to extract all needed weight elements on a per-cycle basis. This bandwidth requirement is dictated both by the number of weights that need to be fetched for each activation and the number of activations that are processed in parallel: 

\begin{enumerate}
    \item \textbf{Weight Sparsity}: for each non-zero activation, the elements from the corresponding compacted dense weight kernels are processed in parallel. As weight sparsity is increased (i.e. smaller $N$), the width of the port required to support the parallel read of all the associated data decreases linearly.
    
    \item \textbf{Activation Sparsity}: processing for each non-zero activation requires an independent lookup. If activations are processed in parallel, each operation requires its own memory port. As activation sparsity is increased (i.e. smaller $K$), the number of memory ports falls linearly.
\end{enumerate}

On FPGAs, memory bandwidth requirements are served by numerous relatively small Tightly Coupled Memories (TCM) that are implemented as Static RAMs (SRAMs). On Xilinx platforms, UltraRAMs (URAMs) are dual-ported, with a port width of 72 bits and a capacity of 288 Kbits (4096 locations). In our implementation, memory requirements for the first few stages of a network such as ResNet-50 will be driven by bandwidth rather than capacity. In order to achieve the stipulated throughput target, weights must be distributed across a larger number of URAMs than would be dictated by storage capacity requirements alone. The memory bandwidth required to support the computation of a 1x1 [64:64] convolution in a single cycle (i.e. fully parallelize [64:64] channel dot products) necessitates multiple dual-ported URAMs. As a result, the storage capacity of each URAM unit is relatively underutilized. 

In summary, in this experiment where we employ a high degree of parallel computation to make an aggressive but fixed throughput target, sufficient local memory bandwidth is key. The pattern of access is not predictable, due to the dynamic selections of the $k$-WTA module. Although this disrupts location access coherence, the rate of access is predictable. This rate is a combined function of both weight and activation sparsity. Compared to an equivalent fully parallel dense network, sparse-sparse networks deliver significant reductions in both the number and capacity of FPGA TCMs, and the associated bandwidth.

\section{Discussion}
Over the last decade there has been significant attention focused on accelerating DNNs using FPGAs and other architectures, including convolutional networks~\cite{orig_cnn}. In this section we compare our approach to research that is closest to our work, discuss some of the issues that arise in deploying our solution to complex networks, and suggest some directions for future work.

\subsection{Accelerating Sparse DNNs on FPGAs}
With Complementary Sparsity we demonstrated that sparse filter kernels can be interleaved such that multiple convolutional kernels are processed simultaneously. A related idea has appeared in \citep{k9} where they compact columns of a weight matrix used in a matrix multiply implementation of convolution, which is then processed through a bit-serial systolic array. As such they can reduce the number of MAC operations by a factor of 8 (see below for additional discussion on systolic arrays). They also discuss a process for creating an interleaved weight matrix by incrementally pruning and compacting during training.  Although they did not explicitly discuss sparse-sparse optimizations, their compaction technique could potentially be adapted for creating complementary sparse kernels that are compatible with our implementation.

There have been a number of papers investigating sparse-dense network implementations on FPGAs. Employing either weight~\citep{k1,k2,k3,k5,k8,k9,k10} or activation sparsity~\cite{k7}, they show it is possible to reduce the number of MAC operations by routing a subset of the dense values to the sparse set of operands at the processing units. This can be done either via multi-ported memories~\citep{k3} or multiplexor networks~\cite{k2}. Although reducing the number of multiplies results in power savings, these techniques typically perform only one dot product at a time in each processing unit. Unlike these methods, Complementary Sparsity makes full use of dense activations and sparse weights. Each activation is paired with a corresponding weight value which allows multiple dot products to be performed every cycle and enables fully parallel operations. In addition, Complementary Sparsity provides a path to sparse-sparse implementations.

\subsection{Accelerating Sparse Networks on Other Platforms}
Recognizing that hardware limitations have held back the deployment of sparse networks~\citep{Hooker2020}, there has been increasing interest in accelerating sparsity on GPU platforms. It is possible to extract meaningful performance gains with block-sparse kernels by implementing large blocks, of size $32\times32$ or larger \citep{Gray2017} with a potential negative impact on accuracy. In \citep{Gale2020} CSR based techniques are used to accelerate common DNN networks such as MobileNet \citep{Sandler2018}. However, the end to end performance gains are limited and restricted to about $1.2\times$ and $2\times$ increase over the dense implementations, respectively. Recently NVIDIA has introduced native support for sparsity in their Ampere~\cite{ampere} architecture. In Ampere there is a limit of 50\% sparsity and end-to-end gains are modest at about $1.3 \times$ faster than dense. To date, GPU based techniques are limited in their ability to achieve significant performance gains on full networks. In addition they do not provide a path to exploiting both sparse activations and sparse weights.

For sparse-sparse networks, when both weight and activation sparsity are employed~\citep{k14,k13,k11,k6,k12,k4}, it is difficult to efficiently pair the non-zero weight and activations.  Many emerging solutions are based around 2D systolic arrays of processing units, on both FPGAs and custom ASIC designs. Here each processing unit performs a check for either matching indices~\citep{k11,k13} or non-zeros~\citep{k4} as the weight and activation values are streamed through the systolic array. One concern with this approach is overall performance.  With Complementary Sparsity we are able to parallelize computation such that we can execute an entire $1\times1$ conv block, representing many sparse kernels, in one cycle. Systolic arrays fundamentally require several cycles to flow through the weights and activations. This process would then have to occur for each sparse kernel, thereby limiting their performance gains.  Finally, implementing them efficiently often requires the costly development of specialized hardware. In contrast, Complementary Sparsity can deliver performance gains today on currently available hardware. 

In this paper we have focused on standard DNNs. Spiking Neural Networks (SNNs) represent an alternate formalism that offers significant potential for performance improvements \citep{Ghosh-Dastidar2009, Roy2019TowardsComputing}. SNNs model neurons using an analog, continuous time, framework. Neurons in SNNs have high temporal sparsity, i.e., they rarely become active. Hardware chips are emerging that exploit this characteristic to create event-based systems that achieve significant energy efficiencies \citep{Davies2018, Roy2019TowardsComputing}. SNNs historically have been unable to match the accuracy of DNNs on complex tasks, an issue that has held back their wide-scale deployment. This problem is an active area of research, with promising recent results~\citep{KIM2021686, TAVANAEI201947}, including approaches that attempt to model the temporal sparsity of SNNs in DNN systems~\citep{DBLP:journals/corr/abs-2107-07305}. 

There exist a number of emerging hardware architectures for exploiting sparsity. In \citep{kendall2020b} the authors review different factors for DNNs, including activation and weight sparsity, and compare a large set of architectures. They suggest that analog crossbar-based architectures represent the most promising direction.  This is also investigated in \citep{Azghadi2020} where they review memristors, memristive crossbars, FPGAs, and SNNs for embedded healthcare applications. Another approach is to implement a scatter-compute-gather module to aggregate operands based upon the indices of their non-zero values~\citep{k14}. In \citep{Zyarah2020} the authors implemented a completely custom memristor-based mixed signal architecture. They demonstrate large performance gains and energy efficiencies for embedded applications using a biologically inspired sparse-sparse learning algorithm. 


\subsection{Deploying Complex Sparse-Sparse Systems}

Our results indicate that it is possible to create convolutional networks that exploit both sparse activations and sparse weights. In this article we presented results for an end-to-end speech network as well as the core components used in most convolutional networks. Although these components can form the foundation for building many networks, modern convolutional networks often contain a large number of layers and a variety of structures. In these networks a number of other issues come into play when designing end-to-end systems. These issues, outlined below, are important design considerations in implementing efficient commercial systems based on Complementary Sparsity.

\emph{Channel Partitioning}: The number of channels associated with the convolutional kernels is not constant and often increases for the deeper layers. For example, in a Resnet-50, layers start with $64$ channels, but this increases to $2048$, as illustrated in Figure~\ref{fig:resnet}. However, as explicitly noted in \citep{He2015a}, the feature map size is reduced correspondingly, keeping the computational requirements roughly constant. In ResNet-50, all convolution operations can be decomposed into groups of 64 dot-products between 64 element vectors, enabling the increasing channel dimension to be handled by the repeated use of our modular [64:64] channel blocks. Our implementation of the $k$-WTA operator also processes the output of the convolutions in units of 64 elements, enabling the modular construction of the ResNet-50 layers.

\emph{Pipeline Latency Balancing}: When balancing the pipeline of an implementation with multiple layers, carefully ``right-sizing'' the layers is important to maximize efficiency and minimize resource utilization. This in particularly important in sparse-sparse networks. As discussed in Section~\ref{sec:stem} we find that the large gains achieved by Complementary Sparsity can lead to unexpected bottlenecks in other areas, such as the initial stem layer. For dense implementations, the main option is a choice between serial or parallel implementations. However, for sparse networks, we also have an additional option. Increasing weight and/or activation sparsity for a given layer translates into reductions in compute operations per layer, reducing (serial) latency, and reducing the memory bandwidth required to supply the operands to the computation. 

\emph{Training Accuracy}: An important issue, outside the focus of this article, is the ability to train sparse-sparse networks that have sufficient accuracy while retaining high sparsity. As discussed in Section~\ref{sec:dnn_sparsity} research in training sparse networks has increased significantly. It is now possible to create accurate networks with 90\% sparsity on ImageNet \citep{pmlr-v119-evci20a} and Transformers \citep{Cohen2021SparsityParameters}. Most of that work has focused on weight sparsity with a few papers focused on activation sparsity. There is relative lack of research on networks that have both forms of sparsity (exceptions are \citep{Ahmad2019,Zyarah2020}). In some scenarios networks trained without explicit activation sparsity end up with highly sparse activations anyway \citep{k2,Javed2019,Beaulieu2020}. This is encouraging because it suggests that sparse activations may naturally be an optimal outcome.  We hope the performance results shown in this article will help lead to additional research on sparse-sparse networks. 

\subsection{Future Directions}
We have presented an initial set of results on Complementary Sparsity, and there are a number of areas for future research.  One direction is to look beyond convolutional networks and apply Complementary Sparsity to other important architectures, such as Transformers \citep{Vaswani2017}, and Deep Recommender systems \citep{zhang2020}. This will require a greater focus on linear layers, where it is possible to overlay multiple rows or columns from a layer's sparse weight matrix. A second promising direction is to leverage our FPGA designs to create hardened IP blocks for a variety of ASICs. A third area is to consider the application of Complementary Sparsity to existing hardware platforms beyond FPGAs\if@blind\else~\cite{why_slow}\fi. Finally, it would be interesting to see if Complementary Sparsity can be used to accelerate the training of sparse-sparse networks. 

\section{Conclusions}
In this article, inspired by the high levels of sparsity in the brain, we investigate the performance benefits of DNNs that exploit both weight and activation sparsity. Using a novel technique that we term \textit{Complementary Sparsity} we show that it enables highly efficient sparse-dense and sparse-sparse networks. Using FPGAs we demonstrate that individual sparse-sparse networks can outperform standard dense DNN networks by over 30X. We further illustrate that sparse-sparse networks can be implemented using far fewer hardware resources than their dense counterparts, and that the resource requirements are inversely proportional to the degree of sparsity. This frugal use of resources allows 5X more networks to be accommodated on an FPGA, delivering a full-chip throughput over 110X higher than the corresponding dense networks. Complementary Sparsity also enables the deployment of DNNs on smaller embedded platforms than previously possible. To our knowledge, we are the first to report such dramatic benefits for both sparse-dense and sparse-sparse networks on FPGAs.

\section*{Acknowledgements}

\if@blind
Acknowledgements removed for blind review
\else
We would like to acknowledge the guidance and feedback from Xilinx, especially Prasun Raha, Vamsi Nalluri, Mrinal Sarmah, and Mary Low. We thank Greg Maltz for his critical feedback and help reviewing and editing the article. Additionally, we would like to acknowledge the help and feedback from Jeff Hawkins, Luiz Scheinkman, Celeste Baranski, Enno Wein and the Instigate team (Naira Khurshudyan, Marine Tumasyan, Hmayak Arzumanyan, Hasmik Mantshyan, Armenuhi Petrosyan, Armen Ohanyan, Davit Petrosyan, Anna Sargsyan, Armen Hovhannisyan, and Tatul Yeghiazaryan) over the course of this research.
\fi

\bibliography{references,subutai_mendeley.bib}
\end{document}